\documentclass[lettersize,journal]{IEEEtran}
\usepackage{amsmath,amsfonts}
\usepackage{algorithmic}
\usepackage{algorithm}
\usepackage{array}
\usepackage[caption=false,font=normalsize,labelfont=sf,textfont=sf]{subfig}
\usepackage{textcomp}
\usepackage{stfloats}
\usepackage{url}
\usepackage{verbatim}
\usepackage{graphicx}
\usepackage{cite}

\usepackage{color}
\usepackage{yfonts}

\begin{document}

\title{Estimating Classification Confidence Using Kernel Densities}




\author{Peter Salamon, David Salamon, V. Adrian Cantu, Michelle An, \\
Tyler Perry,
Robert A. Edwards, and Anca M. Segall

\thanks{Funding: V. Adrian Cantu, Michelle An, Tyler Perry, Rob Edwards, and Anca Segall and some of the computing resources used were supported by a subcontract to Rob Edwards and Anca Segall from a NIDDK grant, RC2DK116713 Computational and Experimental Resources for Virome Analysis in Inflammatory Bowel Disease (CERVAID), to David Wang.}
\thanks{Peter Salamon is with the Department of Mathematics and Statistics and the Computational Science Research Center and the Viral Information Institute, all at San Diego State University, San Diego, CA 92182, USA. (e-mail:psalamon@sdsu.edu)}  
\thanks{David Salamon is with the Department of Mathematics and Statistics at San Diego State University, San Diego, CA 92182, USA. (e-mail:dls@lithp.org)}  
\thanks{V. Adrian Cantu was with the Computational Science Research Center and with the Department of Biology at San Diego State University, San Diego, CA 92182, USA. He is now with the Perelman School of Medicine at the University of Pennsylvania, 425 Johnson Pavilion, 3610 Hamilton Walk, Philadelphia, PA 19104, USA. (e-mail:garbanyo@gmail.com)}
\thanks{Tyler Perry was with the Computational Science Research Center at San Diego State University, San Diego, CA 92182, USA. He is now with Google, 1600 Amphitheater Parkway, Mountain View, CA 94043, USA. (e-mail:tylerjperry16@gmail.com)}
\thanks{Michelle An was with the Bioinformatics and Medical Informatics Program at San Diego State University, San Diego, CA 92182, USA. She is now with Helix, 9875 Towne Center Drive, San Diego, CA 92121, USA. (e-mail:michellean92@gmail.com)}
\thanks{Robert A. Edwards was with the Department of Biology and the Computational Science Research Center and the Viral Information Institute and the Bioinformatics and Medical Informatics Program all at San Diego State University, San Diego, CA 92182, USA. He is now with Flinders Accelerator for Microbiome Exploration, College of Science and Engineering, Flinders University, Flinders, Adelaide, SA, Australia. (e-mail:raedwards@gmail.com)}
\thanks{Anca Segall is with the Department of Biology and the Computational Science Research Center and the Viral Information Institute and the Bioinformatics and Medical Informatics Program, all at San Diego State University, San Diego, CA 92182, USA. (e-mail:asegall@sdsu.edu)}
}
\maketitle

\begin{abstract}
This paper investigates the post-hoc calibration of confidence for “exploratory” machine learning classification problems. The difficulty in these problems stems from the continuing desire to push the boundaries of which categories have enough examples to generalize from when curating datasets, and confusion regarding the validity of those categories. We argue that for such problems the “one-versus-all” approach (top-label calibration) must be used rather than the “calibrate-the-full-response-matrix” approach advocated elsewhere in the literature. We introduce and test four new algorithms designed to handle the idiosyncrasies of category-specific confidence estimation. Chief among these methods is the use of kernel density ratios for confidence calibration including a novel, bulletproof algorithm for choosing the bandwidth. We test our claims and explore the limits of calibration on a bioinformatics application (PhANNs)\cite{cantu_phanns_2020,cantu_phanns_2021} as well as the classic MNIST benchmark\cite{lecun_mnist_1998}. Finally, our analysis argues that post-hoc calibration should always be performed, should be based only on the test dataset, and should be sanity-checked visually.
\end{abstract}


\section{Introduction}
The pressing need for accurate estimators of confidence in classification by machine learning systems has recently been discussed by many authors\cite{gawlikowski_survey_2021,kuppers_multivariate_2020,lakshminarayanan_simple_2017,jiang_trust_2018,zhang_mix-n-match_2020,guo_calibration_2017}. This was mostly in response to Lakshminarayanan et al.’s 2017 NIPS address calling for more attention to the calibration problem\cite{lakshminarayanan_simple_2017,zhang_mix-n-match_2020}. A motivating claim here is that modern training methods have focused on improving accuracy of predictions, often at the cost of calibration\cite{zhang_mix-n-match_2020,guo_calibration_2017,kumar_trainable_2018,kull_beyond_2019,wen_combining_2020}. Miscalibrated predictions can lead to misallocated resources and catastrophic errors. While this is particularly acute for medical or military decisions, good calibration is important for all applications.

In the present manuscript we focus on the post-hoc calibration of the confidence associated with a particular choice by a classifier, where by confidence we mean the probability that the chosen class is correct, and by calibration we mean the rule that associates such probability to the output of the classifier. The motivation for our focus comes from bioinformatic classification problems, but applies to any area in which the class-specific information captured in the dataset differs wildly between the classes. Our approach is predicated on the conviction, noted in PhANNs\cite{cantu_phanns_2020}, that confidence calibration for classification in such problems should be individualized for each class, i.e., confidence assessment should be done separately for each category.

Poorly defined categories with limited examples are typical of “cutting edge” research that deals with exploratory problems. In such fields, machine learning advances the frontiers of knowledge by helping us identify which categories can be objectively identified in the data, for example classes of proteins with distinct functions. Datasets for different protein families do not grow at equal rates, and some functional protein families are less homogeneous than others. In these domains, classifying the shifting and subdividing categories is a daunting task.

\section{Specific applications to bioinformatics and biological datasets}
Multi-class prediction problems in bioinformatics rely on datasets that are parsed and curated from rapidly changing databases using search terms – for example, annotations that may be insufficiently documented or protein functional classes that are often not well delineated. The available datasets for the different classes and the associated reliability of the classifications vary widely. Classifying a protein sequence as belonging to one functional category or another can be very difficult for some classes and relatively easy for others, because some proteins diverge much more than others without losing their function, and because some functional categories have significantly more annotated examples than others. For example, the sequences of the spike protein of influenza or corona viruses frequently diverge to a greater extent from each other compared to the RNA polymerases responsible for replicating the genomes of these viruses. The spike proteins are under greater pressure to diversify in order to escape detection and destruction by the immune system, whereas the RNA polymerases are under greater pressure to conserve their ability to copy the RNA genomes of the viruses. The need to report class-specific performance measures has been clear, at least in bioinformatics applications confronted with highly unbalanced, multi-class problems. Such biological applications depend on databases whose information for some classes far exceeds the information available for other classes\cite{cantu_phanns_2020}. Reporting overall accuracy (or its many relatives, such as F1 scores) is not enough to measure the quality of the prediction in this context. The accuracy prediction must be class-specific. So must confidence prediction.

In the next section, we present our (somewhat unique) view of the calibration problem. To illustrate our arguments, we use PhANNs (https://phanns.com), a bioinformatics classifier for bacteriophage proteins that predicts a score for inclusion of a given protein sequence into one of 11 distinct functional classes\cite{cantu_phanns_2020}. The score identifying the most likely predicted class is accompanied by a confidence score that estimates the probability that the predicted category is correct. The PhANNs example provides a wide spectrum of categorical behaviors and illustrates our point that different classes need separate consideration for the calibration of confidence. The second example is the classic MNIST dataset of handwritten digits. It was chosen and trained to provide a comparable example that had two independent test sets and could be interrogated by a statistically significant sample of ensembles of ten networks, all trained with 30,000 examples. The Appendix gives further details about the training and results for these networks.

Both examples use an ensemble of ten neural networks.The PhANNs scores come from the sum of the strengths of the separately trained networks’ softmax outputs. This results in PhANNs scores being on a 0-10 scale rather than the more customary 0-1 scale. The temptation to divide by ten was resisted on the grounds that the score would be mistaken for a calibrated confidence, which for some categories it certainly was not. For the MNIST scores we used the more usual 0-1 scale.

\section{Confidence in Category Prediction}
Confidence is a measure of our trust in the category predicted by the machine learning system (MLS). For virtually every author on the subject, this measure is the probability that the prediction is correct. The post-hoc calibration task is to estimate this probability given the post-training MLS and a test set. What information to use in estimating this probability is a matter of some debate. In particular, what information from the MLS is assumed to be known in calculating the conditional probability is (surprisingly) an issue. In a recent paper, Gupta and Ramdas\cite{gupta_top-label_2022} argue that the category predicted by the MLS should always be part of the given information when associating a confidence to the score. We wholeheartedly agree with their arguments, and have argued for the same category-specific approach (top-label calibration) based on independent reasoning having to do with poorly and heterogeneously defined categories leading to noisy and unbalanced datasets. 

Figure 1 shows the joint distribution of true category and predicted category in the PhANNs test set of 46801 samples. The Figure also shows two normalizations of the joint distribution into conditional distributions, normalized either by row or by column, i.e. conditioned on the true category versus conditioned on the predicted category. The columns of the table on the bottom right give the confidence at this level of information, i.e. conditioned only on the category predicted. In most cases, the diagonal entry in the Reverse Confusion Matrix, i.e. the confidence, is worse, sometimes much worse, than the corresponding diagonal entry in the Confusion Matrix, i.e. the true positive rate. A particularly egregious example is the confidence in the prediction of Minor Capsid; the conditional probability that a sample identified as a Minor Capsid is indeed a Minor Capsid, is 0.07. 
\begin{figure}[!t]
\centering
\includegraphics[width=3.5in]{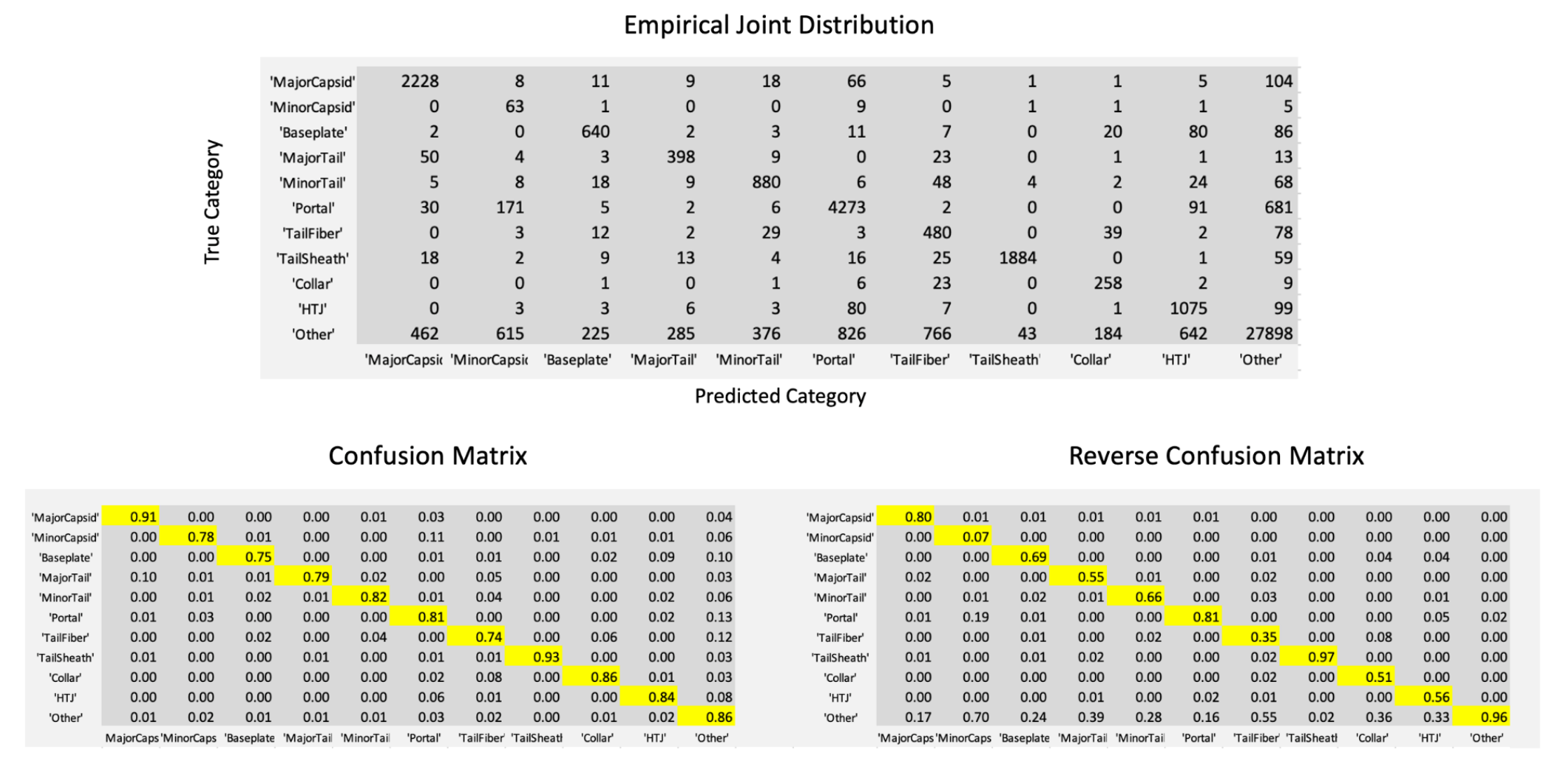}
\caption{The empirical  joint distribution of true and predicted categories in PhANNs. The lower left panel shows the data normalized by row (each entry divided by the sum of the entries in its row). The resulting entries represent empirical estimates of the probabilities of each predicted category given the true category. The lower right panel is created similarly, this time normalizing by columns. The resulting entries represent empirical estimates of the probabilities of each true category given the predicted category. The diagonal entries of the Reverse Confusion Matrix are the confidence Conf1 in equation (1).}
\label{fig_1}
\end{figure}

Figure 2 shows the three levels of output from an MLS trained to perform classification. Given an input vector $X$, the MLS always predicts the category $k=k(X)$ assigned to $X$. It may also output a confidence score $S(X)$ or even a probability distribution $Y(X)$ over the categories. This leads to three possible interpretations of \textbf{confidence in the predicted category}, depending on the level of information given in the conditions. The first of these is just
\begin{equation}
     \text{Conf1} (X) =  
    \text{Prob}\Big(\text{TrueClass}(X)=k(X) \, \Big| \, k(X)\Big) , 
        \label{Conf1}
\end{equation}  
the empirical estimates of which appear as the diagonal entries of the reverse confusion matrix in Figure 1. 
\begin{figure}[!t]
\centering
\includegraphics[width=3.5in]{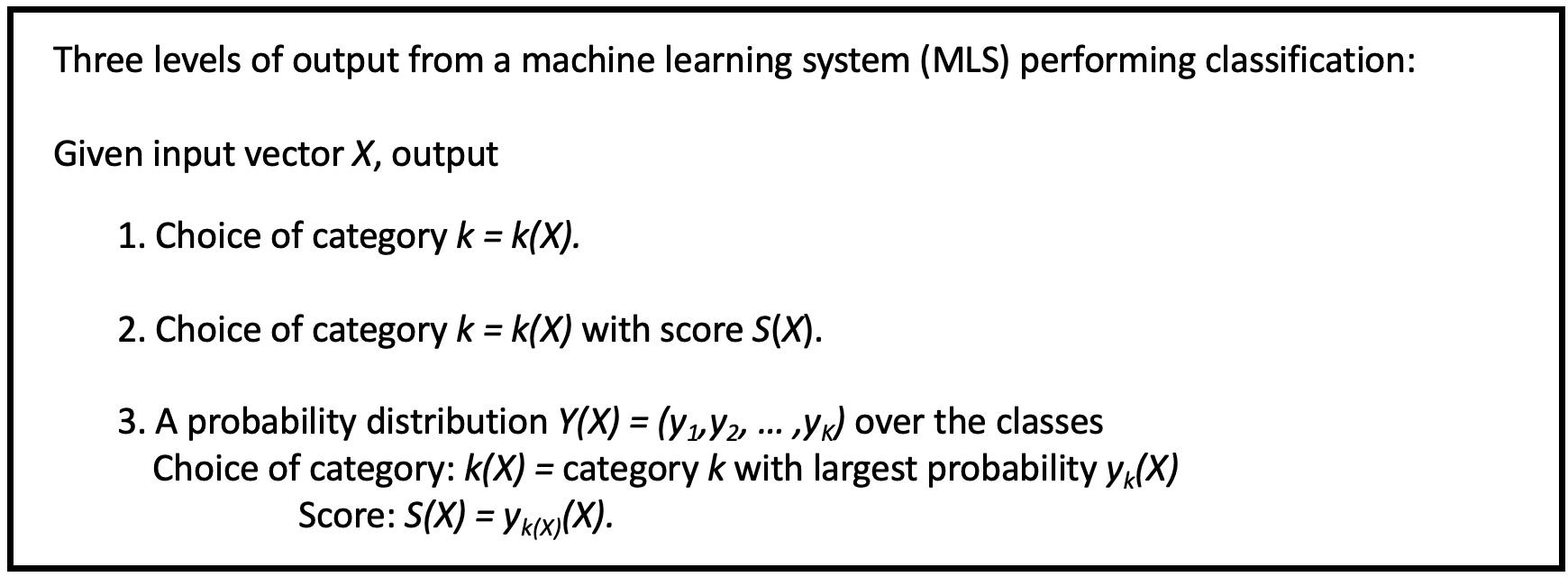}
\caption{Schematic division of MLS into three groups based on information output from the MLS.}
\label{fig_2}
\end{figure}

When we have more information in the form of a score function, the corresponding definition of confidence becomes
\begin{eqnarray}
\nonumber
 \text{Conf2} (X) & = & \text{Prob} 
\Big(\text{TrueClass}(X)=k(X) \, \Big| \, k(X), S(X) \Big)    \\
& = & \text{Conf}_k(S) 
\label{Conf2}
\end{eqnarray}

This confidence is our primary focus and the one that is the most natural within the top-label paradigm. Calibration at this level is the estimation of the probability density in equation (\ref{Conf2}) as a function of score and comes in the form of the $K$ functions Conf$_k, k=1,...K$, that convert scores $S$ on samples having $k(x)=k$ to probabilities of being correctly classified. Note that this divides the confidence estimation portion of the multi-classification problem into $K$ separate one-versus-all problems\cite{kull_beyond_2019,gupta_top-label_2022,zadrozny_transforming_2002}. We discuss our recommended implementations for finding good estimators of the functions Conf$_k(S)$ in the following sections.

The score function $S(X)$ above can come from several possible sources. Usually it is the (uncalibrated) confidence that came with the classification, and reflects performance on the training and/or validation set. For random forests, it is typically the fraction of trees voting for this classification. For MLS type 3 (see Figure 2) the $k$-th component of the output vector $Y(X)$ is supposed to be the probability that the input $X$ belongs to category $k$. The entry for the chosen category, $y_{k(X)}(X)$,  should thus be the confidence we seek. For post-hoc calibration, this entry in $Y$ can serve as the score $S(X)=y_{k(X)}(X)$, and is worth recalibrating\cite{zhang_mix-n-match_2020,guo_calibration_2017}, especially in light of post-training dataset growth\cite{ovadia_can_2019}. 

While the most important entry in the MLS output $Y(X)=(y1, \ldots, yK)$ for deciding confidence calibration is certainly $y_{k(X)}$,  the other entries also carry some information. We explore some calibration schemes using the full K-tuple $Y(X)$ in the discussion of parametric models below. Notably, we introduce some top-label oriented extensions to temperature scaling.

\section{Histograms}
The earliest confidence calibration methods\cite{zadrozny_transforming_2002,drish_obtaining_2001,platt_probabilistic_1999} are called histogram methods. They work by binning the scores and defining the confidence for a bin using the empirical fraction of samples whose predicted category matches the true category. The procedure is illustrated in Figure 3.
\begin{figure}[!t]
\centering
\includegraphics[width=3.5in]{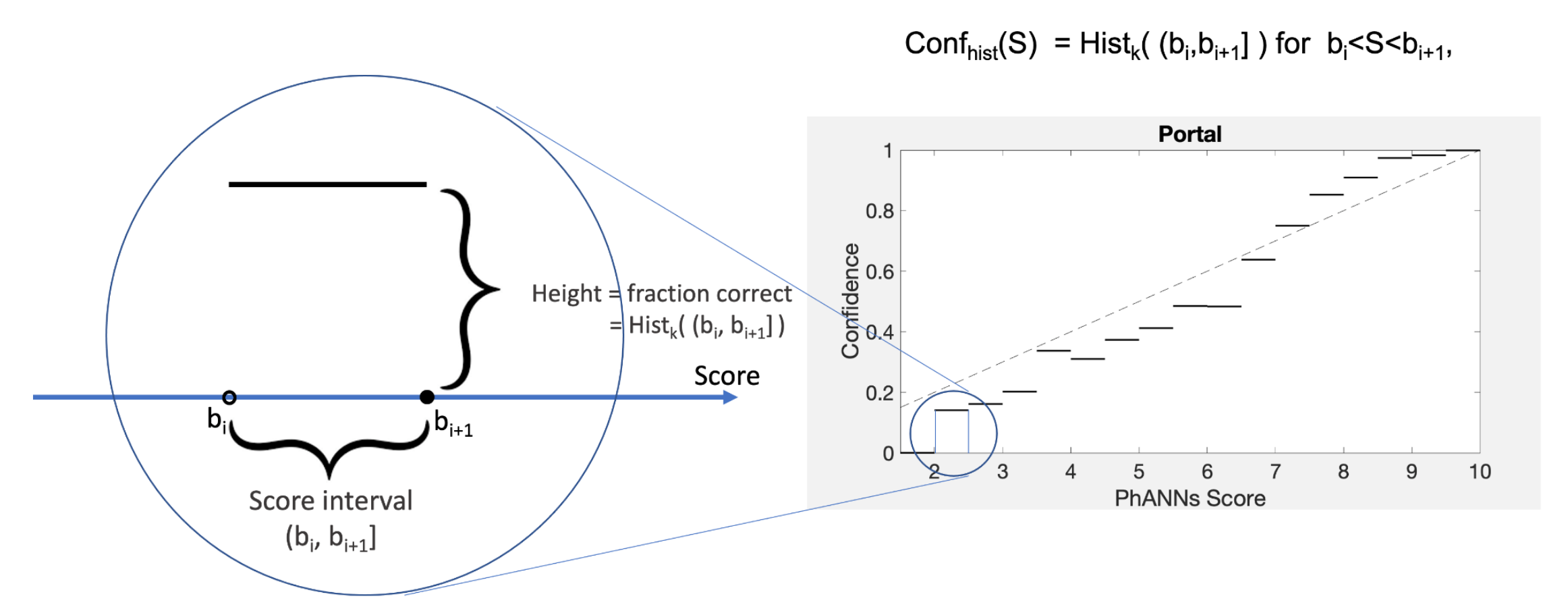}
\caption{The fraction of true positive samples in a subinterval of the scores range is an estimate of the confidence for that score range. Combining these over a partition of the possible score values creates the graph of Conf$_{\text{Hist},k}(S)$. }
\label{fig_3}
\end{figure}

Following the dictates of our top-label approach, we pay attention to only those samples $X$ in the test set for which the predicted category is fixed, $k(X)=k_0$. We call these samples positive, and divide them into two subsets, the true positives
\begin{equation}
    \text{TP} = \{X; \text{TrueClass}(X) = k_0, \text{and}\,\, k(X) = k_0\}
\end{equation}
and the false positives
\begin{equation}
    \text{FP} = \{X; \text{TrueClass}(X) \neq k_0, \text{and} \,\, k(X) = k_0\}
\end{equation}

Focusing then on an interval of score values, $a<S\leq b$, we can approximate the true confidence for scores in this interval using the empirical estimate
\begin{equation}
    \rm{Hist}_{k_0}\Big((a,b]\Big) = 
\frac{\Big|TP\cap(a,b]\Big|}{\Big|TP\cap(a,b]\Big| + \Big|FP\cap(a,b]\Big|}
\end{equation}
where $|\{ . \}|$ denotes the number of elements in a set. In the formula above, the numerator counts the number of true positive samples with scores in $(a,b]$, while the denominator counts all positive samples with scores in $(a,b]$. The ratio, Hist$_{k_0}((a,b])$, is an average confidence for scores in the interval $(a,b]$. It is the histogram of “histogram calibration” although it is not a histogram in the usual sense. We can piece together a histogram-based Conf function using the values given by Hist$_{k_0}$. Formally, consider a partition $B=\{0=b_1<b_2<...<b_{n-1}<b_n=1\}$ of the interval of possible scores. We then define
\begin{equation}
    \text{Conf}_{\text{Hist}_k} (S)  = \text{Hist}_k\Big((b_i,b_{i+1}]\Big)	\,\, \text{ for } \,\, b_i<S<b_{i+1},
\end{equation}
by combining histograms on each subinterval as illustrated in Figure 3. We use half-open intervals, so their union covers all values of $S$ exactly once.

Using observed frequencies in the test set to make confidence estimates specific to each bin can spread even a respectably sized test set thin. This is what makes the binning approach rather noisy. But as far as giving a noisy picture of “true confidence,” it is best practice. General bounds on the level of noise from arbitrary distributions have been recently established\cite{gupta_distribution-free_2020,gupta_distribution-free_2021}. These bounds are certainly the route to the eventual, high-accuracy confidence values needed for some applications. Gupta and Ramdas’s formalism aimed at much larger datasets than what can be expected to be available for the sort of exploratory problems our PhANNs dataset represents (and which we synthetically emulate with the MNIST dataset). For our target data size range of about 1000 samples from each category, their formalism bounds the histogram error for 10 bins with an expected calibration error under 0.18, ninety percent of the time (see Figure 2 of ref Gupta and Ramdas\cite{gupta_distribution-free_2021}). For our actual dataset sizes in PhANNs, the noise level is even greater since several of the categories have far fewer than 1000 samples. The MNIST dataset, trained to match PhANNs’ performance and unimpeded by balance issues, can provide an easily accessible picture of the noise level. Since the MNIST data effectively comes with two independent test sets, we can draw two independent sets of histograms for the same MLS. Shown in Figure 4 are two sets of empirical ConfHist curves for each MNIST category using ten equal size bins. Their differences show the noise level of empirical estimates for the size range of our examples. For MNIST, each test set consisted of 10,000 examples. For PhANNs, there were 46,000 examples in the test set but almost 30,000 of these belonged to the category “Other”.
\begin{figure}[!t]
\centering
\includegraphics[width=3.5in]{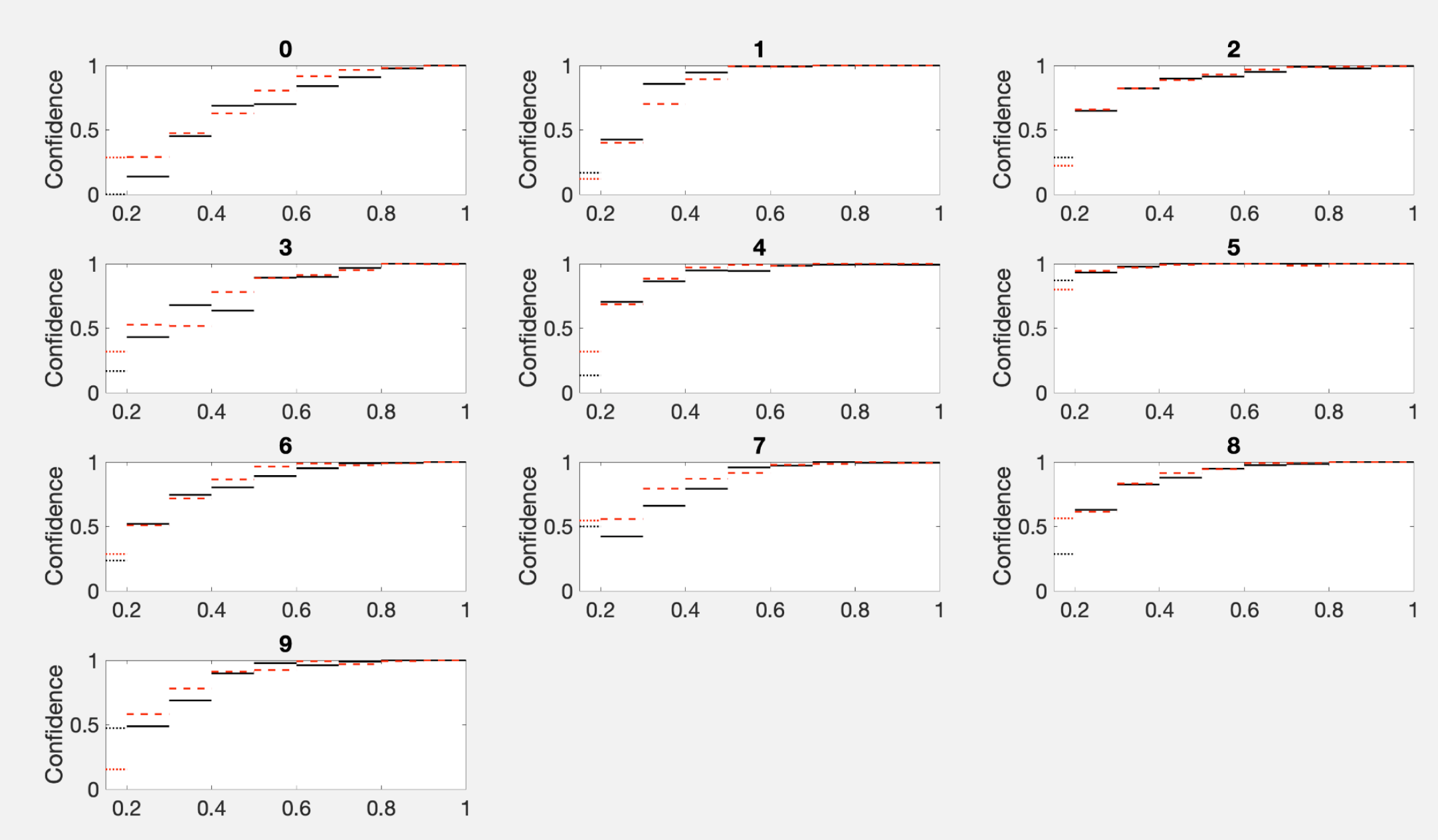}
\caption{Two sets of histogrammed confidence for the MNIST problem (solid black vs dashed red) for each MNIST category using the two independent test sets. The differences between red and black histogram heights are entirely due to noise, showing the limitations of the histogram calibration when only 1,000 test set examples are available.}
\label{fig_4}
\end{figure}

Despite the evident noise level shown in Figure 4, histograms usually achieve among the lowest error rates. In fact, the histogram approach to confidence calibration has found new proponents\cite{gupta_distribution-free_2021}  who show that it has impressive performance. As an anchor, we include a plot of Conf$_{\rm{Hist}}(S)$ using 10 equal sized bins for comparison with all our calibration curves. 

Histograms are considered a non-parametric method. Our favorite approach, described next, is closely related; it is also non-parametric, but avoids binning. 

\section{Kernel Densities}
Our recommended approach to estimating the correctness probability at a specific score is to use the density of similar scores among true positive samples as a fraction of the density of all positive samples. The two densities are estimated in an approach called kernel density estimation, in which the sample points are fattened to small normal distributions which sum to an approximation of the distribution that gave rise to the sample. One could think of it as viewing our sample “slightly out of focus” so each sample point instead appears as a small blur. The choice of the extent of blurring b (known here as the bandwidth) can have a significant effect; we discuss bandwidth selection in section 12 below. 

Specifically, to find the confidence associated with a prediction $k(X)=k_0$, we again pay attention only to those elements of the test set that had predicted category $k_0$. We refer to these samples as positive and recall the definitions of the true positive set, TP, and the false positive set, FP. We then estimate a kernel density for the list of scores in TP 
\begin{equation}
    {\cal K}_{\rm{TP}}(S) = \frac{1}{ |\rm{TP}|} \sum_{X \in \rm{TP}} \frac{1}{\sqrt{2\pi} b} \exp\big(-\frac{(S-S(X))^2}{2b^2}\big)
\end{equation}
and a kernel density for the list of scores in FP
\begin{equation}
    {\cal K}_{\rm{FP}}(S) = \frac{1}{ |\rm{FP}|} \sum_{X \in \rm{TP}} \frac{1}{\sqrt{2\pi} b} \exp\big(-\frac{(S-S(X))^2}{2b^2}\big)
\end{equation}
 We then use these kernels to predict the confidence at any score S as
\begin{equation}
    \rm{Conf}_{\rm{KDE}}(S) = \frac {\cal{K}_{\rm{TP}}(S)\cdot |\rm{TP}| } { \cal{K}_{\rm{TP}}(S)\cdot |\rm{TP}| + \cal{K}_{\rm{FP}}(S)\cdot |\rm{FP}| }
\end{equation}
The procedure is illustrated in Figure 5 for the Baseplate class. 
\begin{figure}[!t]
\centering
\includegraphics[width=3.5in]{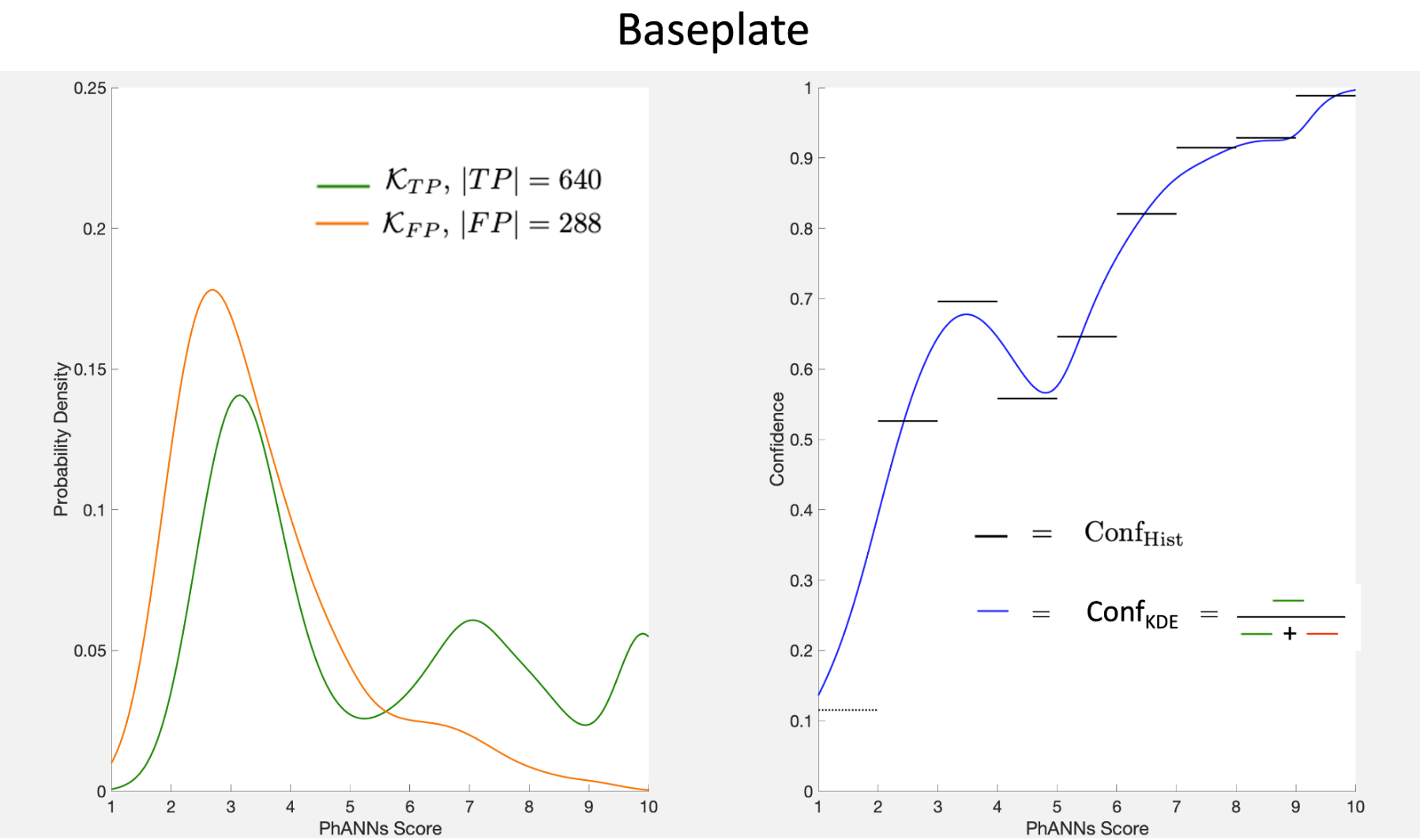}
\caption{Left Panel: Kernel densities for the probability density of finding a sample with a given score among the TruePositive (green) or the FalsePositive (orange) samples. Right Panel: Confidence from the ratio of kernel density estimates (blue) calculated from equation (9) compared to the histogram binned confidence (black).}
\label{fig_5}
\end{figure}

Our approach using the kernel density ratio, Conf$_{\rm{KDE}}$ in equation (9), is a smoothed alternative to the histograms. We are not the first to suggest their use. Zhang et. al.\cite{zhang_mix-n-match_2020} recommend evaluating the calibration error using kernel density estimators (KDEs) for the unknown distributions to enable stable evaluation of the integrals involved. They even recommend integrated deviation from their KDE estimators as the way to estimate calibration error. What they do not seem to realize is that their KDE estimate is in fact a great calibrator. They also miss a crucially important restriction forced by the category predicted. This is the topic of our next section.

\section{Top-Label confidence}
When we say that confidence is the fraction of samples that are correct, we are referring to something like equation (5). The numerator of that fraction is clear but the denominator could be imagined without the $k(X)=k$ condition. In fact Zhang et al.\cite{zhang_mix-n-match_2020} use kernel denominators that do not impose the predicted category condition. 

As we have stressed above, our perspective forces us to focus only on the positive examples, i.e. the ones with $k(X)=k$. There is much information in the predicted category and it should be exploited. Furthermore, it is information that we always have. The right panel in Figure 4 shows the effect of counting all scores on category $k$ (red curve) compared with counting just the scores of samples that predicted category $k$ (blue curve). As the picture makes clear, it is better to count only with the positive scores. Note that scores above 5 automatically forced the sample to predict category $k$ and so the two ways of counting only diverge for scores below 5.

\begin{figure}[!t]
\centering
\includegraphics[width=3.5in]{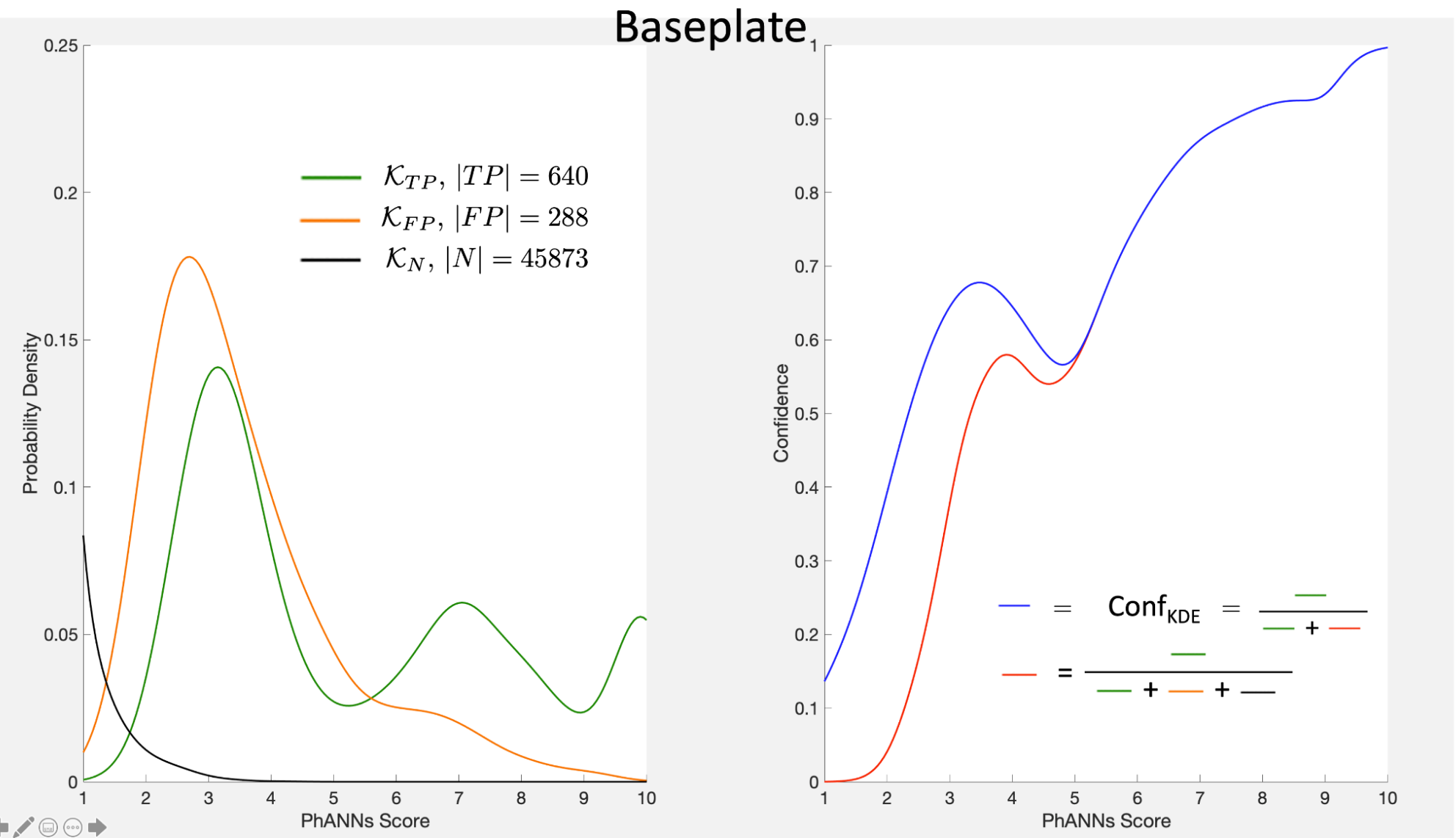}
\caption{The effect of counting scores that did not predict the category. Left Panel: Kernel densities for the probability density of finding a sample with a given score among the TruePositive (green), the FalsePositive (orange), and the score on this category by negative samples (black). Note the much higher weight attached to the negative samples, since most samples did not predict this category. Right Panel: Confidence from the ratio of kernel density estimates (blue) compared with Confidence that would result if all the negative samples were used in the denominator (red). The lesson is to only use those yi with i=k(X) rather than using a diluted signal.}
\label{fig_6}
\end{figure}

This brings us to our last version of the definition of confidence appropriate for the third type of MLS in Figure 2, one whose output $Y(X)=(y_1,y_2,...,y_K)$ is in fact a probability distribution over the classes. 
\begin{equation}
    \text{Conf3} (X)  =  \text{Prob}\Big(\text{TrueClass}(X)=k(X) \,\Big| \, k(X), Y(X) \Big)
\end{equation}

Note that we have stressed that the predicted category, $k(X)$, be part of the given conditions. The entries in the output probability vector $Y(X)$ are supposed to be the probabilities for each category. Thus $y_{k(X)}$ is the (uncalibrated) estimate of the probability we seek. But is it the value on the red or the blue curve above?  \textbf{In fact standard training does not take into account the required difference between the red and blue curves in Figure 4 based on slight changes in activations.}

We recommend using the score function $S(X)=y_{k(X)}$ and calibrating with Conf$_{\rm{KDE}}$ as above. Note, that this does not use any of the entries in $Y$ other than $y_{k(X)}$.  We also discuss some parametric alternatives below that use the full $Y$ vector. Either way, recalibration is worthwhile and cannot hurt. While the uncalibrated confidence estimates $y_{k(X)}$ may in fact be good estimates of the confidence, recalibration will show this to be the case. If the data was calibrated to begin with, a final calibration finds a calibration function Conf$_k$ very close to the identity function. Two examples are shown in Figure 7 for the PhANNs categories Minor Tail and Portal. The calibration function Conf$(S) = S/10$ (the effective identity here since PhANNs scores run 0-10) comes very close in these examples to the results of calibrating using kernel densities or histograms or temperature scaling with awards (see below).

\begin{figure}[!t]
\centering
\includegraphics[width=3.5in]{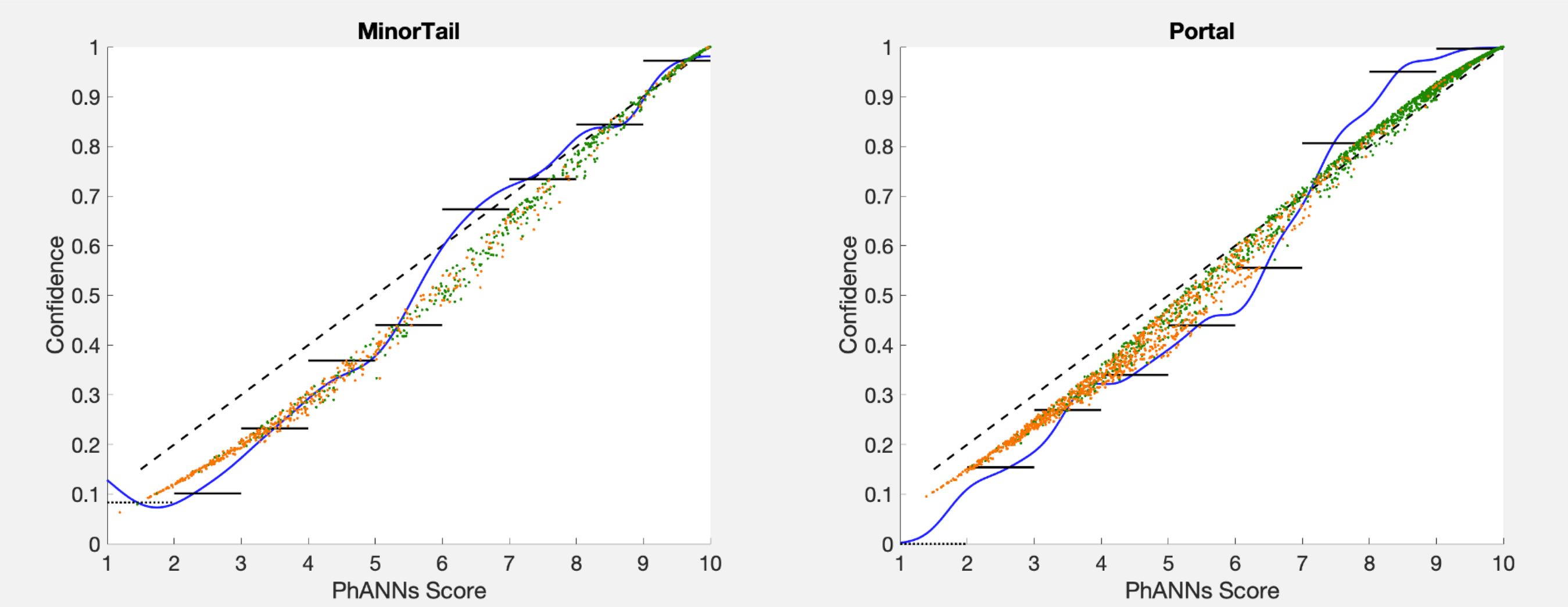}
\caption{The scoring function PhANNs Score / 10 is already very close to calibrated for the categories Minor Tail and Portal. This can be seen from the proximity of the uncalibrated score (dashed line) to our three best confidence estimators: (1) the kernel density estimator Conf$_{\rm{KDE}}$ (blue), (2) the histogram estimator Conf$_{\rm{Hist}}$ (black), and (3) the cloud of estimates from temperature scaling with awards (green and orange), described in detail in section 11 (TSwA).}
\label{fig_7}
\end{figure}

\section{The PhANNs Approach Tweaked}
The original PhANNs paper assigned confidence values in category $k$ by associating to a sample $X$ with score $S(X)$ the empirical fraction Hist$_k([S(X),\infty])$. This corresponds to the very liberal strategy of grouping a score $S$ with all scores greater than or equal to $S$. We shall refer to it as the cumulative confidence Conf$_{\rm{Cum}}$. 
\begin{equation}
    \rm{Conf}_{\rm{Cum}}(S) = \rm{Hist}_k\Big([S,\infty]\Big)	
\end{equation}

Using cumulative distributions to mitigate the noisy empirical estimates of small bins is less sensitive to sampling noise than binning strategies. The approach is also very robust, i.e. it will work on examples where the other methods do not. It does require a careful statement warning the user that the returned value is the probability of a score of S or better, rather than the probability density at S. Here we discuss adapting Conf$_{\rm{Cum}}$ as an estimator of the probability density.

The cumulative features imply that in general Conf$_{\rm{Cum}}$ is much slower to come down toward lower estimates for lower scores. This is illustrated in the top left panel of Figure 8 for the category Tail Fiber. Not surprisingly, this confidence estimator badly overestimates the confidence for low scores. A simple trick however improves this estimator significantly.
\begin{equation}
    \text{Conf}_{\text{Cum+}}(S) = \begin{cases}
  \text{Conf}_{\text{Cum}}(S) & \text{if } S\geq\theta \\
  \text{Hist}_k([0,\theta]) & \text{if } S<\theta
   \end{cases}
\end{equation}

The confidence function Conf$_{\rm{Cum}+}$ uses Conf$_{\rm{Cum}}$ from the max score down to score $S=\theta$, below which it uses just one big histogram bin. If we choose $\theta$ equal to the median of all scores that go with category choice $k$, the technique remains non-parametric. The result of doing this is shown in the top right panel of Figure 8. While it makes the method more competitive, it is still far worse than kernel density based estimates or histogram estimates. 

\begin{figure}[!t]
\centering
\includegraphics[width=3.5in]{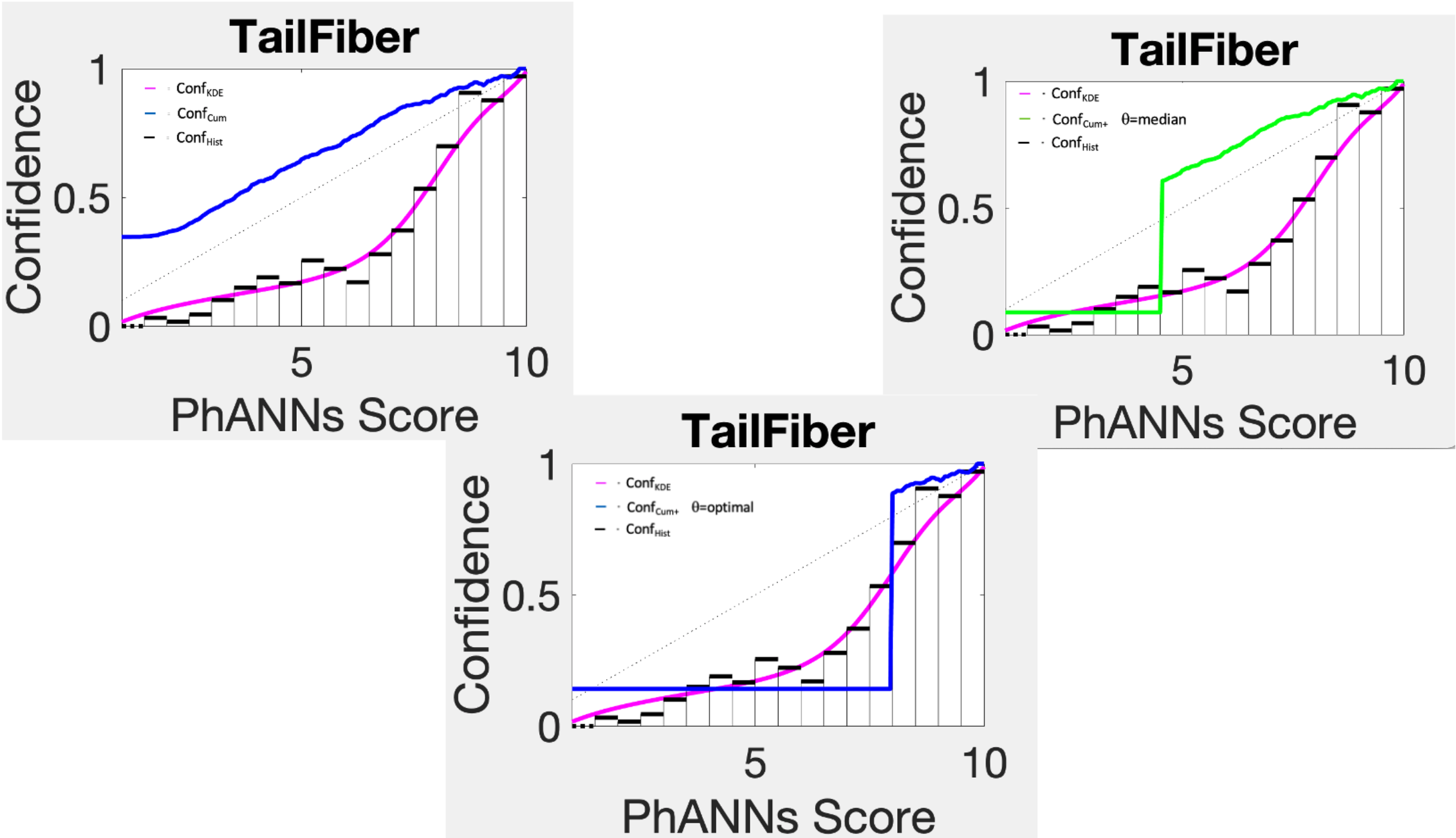}
\caption{Top left panel shows cumulative confidence ConfCum compared to KDE and histogram confidence estimates.  The other two panels compare ConfCum+ using the median as the cutoff (top right panel) and the optimal cutoff (bottom panel).}
\label{fig_8}
\end{figure}

\begin{figure}[!t]
\centering
\includegraphics[width=3.5in]{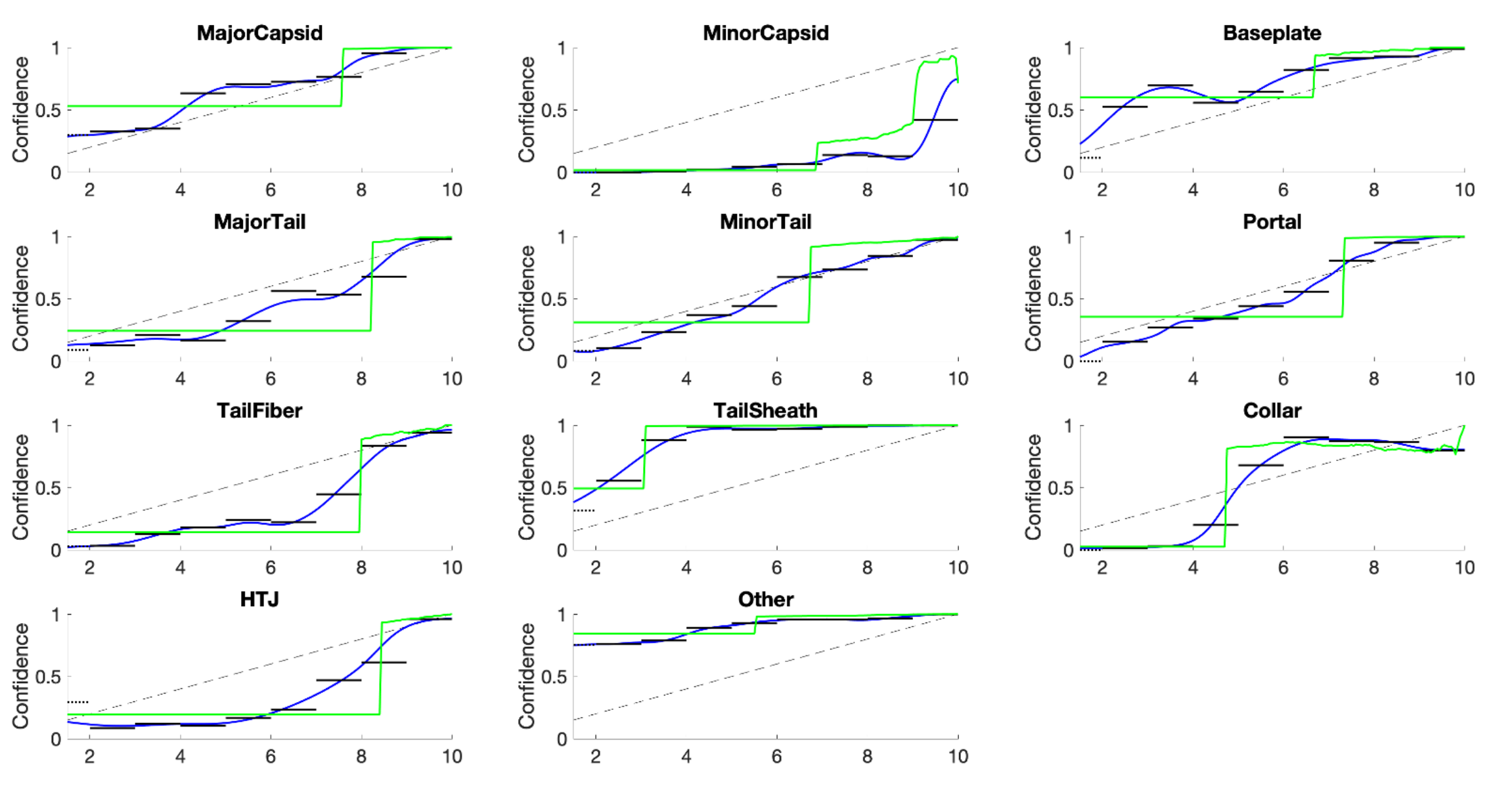}
\caption{Optimally chopped cumulative confidence, Conf$_{\rm{Cum*}}$ (green) compared to Conf$_{\rm{Hist}}$ (black) and Conf$_{\rm{KDE}}$ (blue) on PhANNs categories.}
\label{fig_9}
\end{figure}

With an optimized cutoff $\theta_{\rm{opt}}$, chosen to minimize negative log likelihood in calibration, the resulting confidence function 
\begin{equation}
    \text{Conf}_{\text{Cum}*}(S) = \begin{cases}
  \text{Conf}_{\text{Cum}}(S) & \text{if } S\geq\theta_{\text{opt}} \\
  \text{Hist}_k([0,\theta_{\text{opt}}]) & \text{if } S<\theta_{\text{opt}}
   \end{cases}
\end{equation}

becomes competitive, but not stellar (see Figure 9). This is further corroborated by the charts comparing the performance of our methods on the PhANNs dataset in section 14 where we refer to this method using the acronym CwOC, cumulative confidence with optimal cutoff. 

On the other hand, an optimized cutoff also moves this method into the category of parametric models, where the competition is stiffer. This is the topic of the next several sections.

\section{Parametric Models}
Parametric models involve the selection of a parameter based on data. If that data involves the test set, any predictions using the parameter (such as confidence predictions) face a bias-variance tradeoff. There are four possible approaches – all of them used. 

The first approach basically views the calibration problem as a second machine learning task that therefore needs a second test set\cite{zhang_mix-n-match_2020,kumar_trainable_2018} and can be attacked by elaborate methods. For the exploratory problems we are considering here, this is unlikely to be a viable approach due to paucity of data. 

The second approach is to use the validation set to train these parameters, after the network parameters are fixed. In that case the resulting calibrations are part of the MLS output and should in turn be calibrated post-hoc on the test set by the techniques described above.

A third approach is to use cross-validation on test data. This requires more work but can extract nearly all of the information in the test set with minimal bias. It tends to be quite unstable with sparse data, which is the case for several regions of several categories in the PhANNs dataset. 

The fourth approach is to just train on the test set.  This is our approach here and is the only truly post-hoc approach in the presence of scarce data. With only a few, sufficiently generic parameters, there is little risk of overtraining. Is it a valid confidence calibrator? Certainly. Can we make an unbiased prediction of how well it will perform on other data? Alas, no. For that, we need a second test set. Using the fitted parameters to estimate the error on the same set that was used for the estimate tends to underestimate such error. Here we used experiments with the MNIST dataset to assess the magnitude of the resulting bias, which could not be distinguished from noise for our models below. This is detailed further in Appendix A.

To train our parameters, we followed the procedure adopted by Ovadia et al. and minimized training error on the test set\cite{ovadia_can_2019}. Initially, we trained using NLL loss and Brier score. These are proper scoring functions\cite{gneiting_strictly_2007} so the results are acceptable, but slightly better results are obtained by using NLL or Brier loss following conversion to the effective binary problem for the category following recommended procedure for top-label calibration\cite{gupta_top-label_2022}. Specifically, we used the following loss
\begin{equation}
    \text{Loss}_{\text{NLL}} = - \sum_{X \in TP} \log (S(X)) - \sum_{X \in FP} \log (1-S(X)) 
\end{equation}
which gave results very close to what one obtains using
\begin{equation}
    \text{Loss}_{\text{Brier}} = \sum_{X \in TP} (1-S(X))^2 + \sum_{X \in FP} S(X)^2
\end{equation}
where TP and FP are the category specific true positive and false positive samples as defined above. 

Temperature scaling is a popular parametric technique that estimates one parameter, a temperature T, to (retroactively) adjust the harshness of the softmax used to get $Y$. Introduced by Guo et al\cite{guo_calibration_2017}, it has been positively reported on by several research groups \cite{zhang_mix-n-match_2020,kumar_trainable_2018,kull_beyond_2019,tomani_parameterized_2021}. Our interest here is not temperature scaling per se, but rather some tweaks on temperature scaling that work very well for top-label calibration. Nonetheless, we begin with the standard version.

\section{Temperature Scaling}

Specifically, temperature scaling remaps the MLS output $Y=(y_1,y_2,...,y_K)$ to 
\begin{equation}
    \hat{Y}=\text{softmax}(\frac{\log(y_1)}{T},\frac{\log(y_2)}{T}, ... \,, \frac{\log(y_K)}{T})
\end{equation}

where the temperature T is chosen to optimize the fit to the data using a proper scoring function, exactly like when one is training the MLS. Note that this makes
\begin{equation}
    \hat{y}_i= \frac{y_i^{1/T}}{\sum_{k=1}^K y_i^{1/T}}
\end{equation}

Note that temperature scaling alters all the scores the same way for all categories with one parameter. While this works well for the balanced MNIST dataset where all of the curves are quite similar, it does not work nearly as well for the PhANNs dataset where some of the categories need a temperature greater than one (concave down) and some need a temperature less than one (concave up). This makes temperature scaling a benefit for some categories but a harm to others. In Figure 10 we show the results of temperature scaling for both the PhANNs and the MNIST datasets. 

\begin{figure}[!t]
\centering
\includegraphics[width=3.5in]{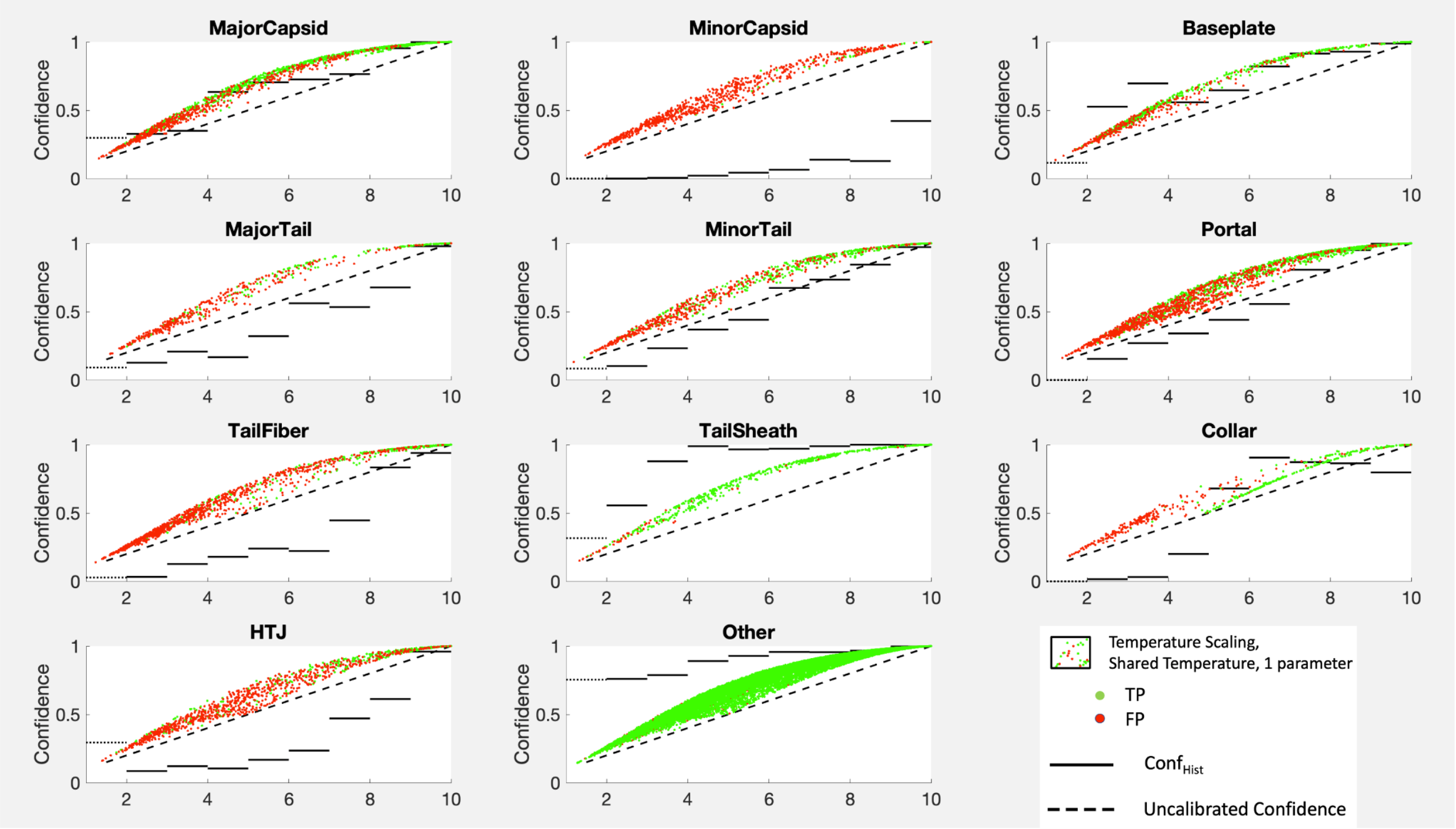}
\includegraphics[width=3.5in]{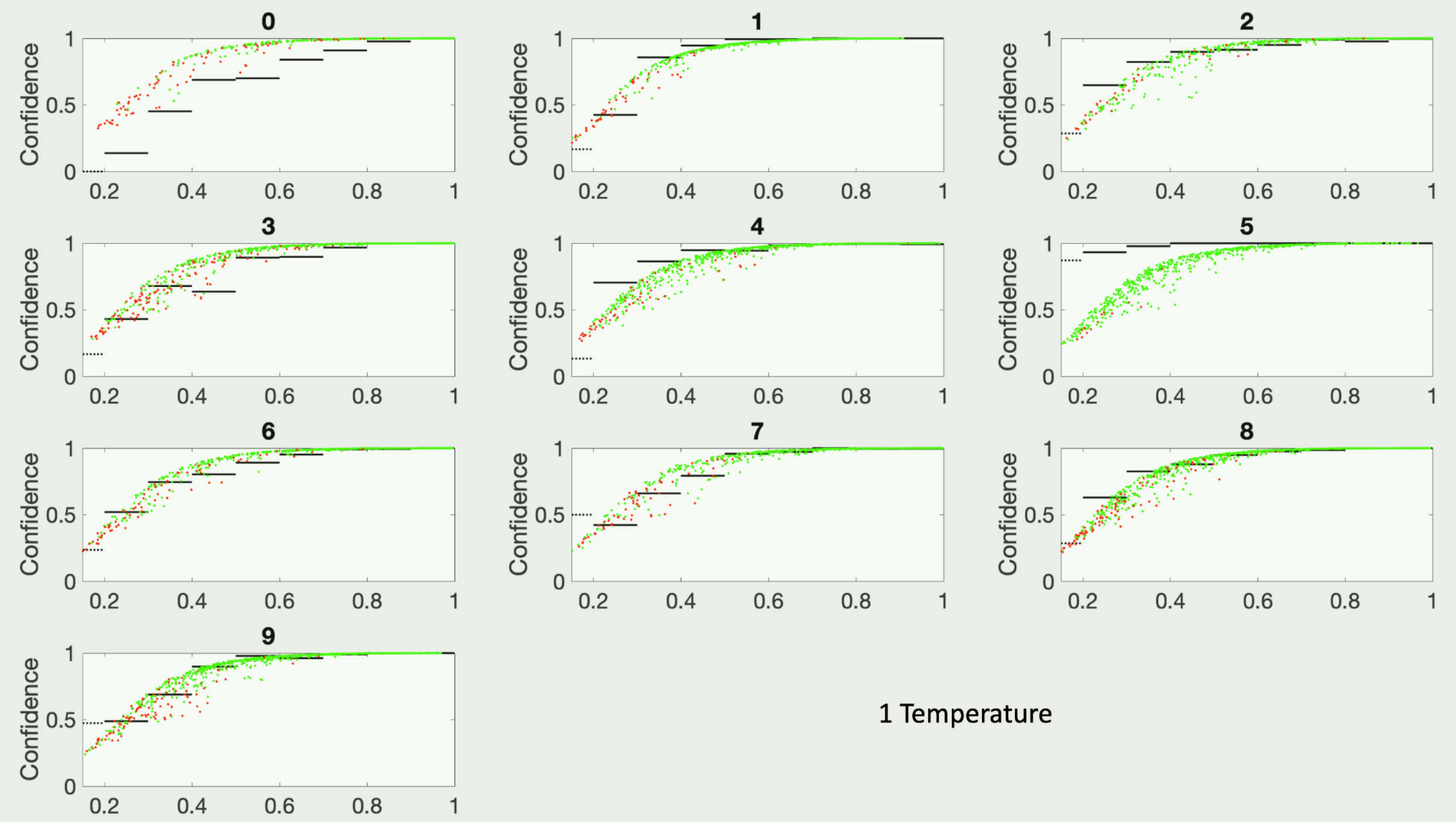}
\caption{Shared temperature scaling for PhANNs in the top panel and for MNIST in the bottom panel. Note that at each score there is a range of corresponding temperature scaled confidence values. Also note that while shared temperature scaling does quite well for MNIST, several categories in PhANNs, e.g. TailFiber and HTJ, would be better fit with the reversed convexity. In each subplot, the horizontal axes show the score while the vertical axes show confidence measured as $y_k$.}
\label{fig_10}
\end{figure}

The cloud of points results because temperature scaling sends a certain score $y_k$ to different $\hat{y}_k$ depending on how the remaining $(1-y_k)$ fraction of the probability is split among the other categories. The Figure shows movement (from the diagonal) generally in the direction of the right answer, shown here by the Conf$_{\rm{Hist}}$ lines.

While we see that temperature scaling improved the calibration of the Other category, it was at the cost of worse calibration on categories such as “HTJ”, which would have benefitted from a temperature greater than one moving all the points downward. Unfortunately, one parameter to recalibrate the full $K$ dimensional output is asking too much, especially when the different categories pull in different directions, as is the case for PhANNs. The PhANNs dataset was dominated by the very populous category “Other” and hence the irresistible pull toward $T<1$. Since the algorithm only gets to choose one temperature for all categories, temperature scaling only goes so far. 

\section{Category Specific Temperature Scaling (CSTS)}
One obvious modification given our philosophy of top-label calibration is to use $K$ different temperatures. The loss function must be adjusted, however, to count only those examples with $k(x)=k$ using the correct answer to be 0 or 1. The result is a $K$-parameter model that is surprisingly competitive with KDEs. The resulting confience curves are shown in Figure 11.

\begin{figure}[!t]
\centering
\includegraphics[width=3.5in]{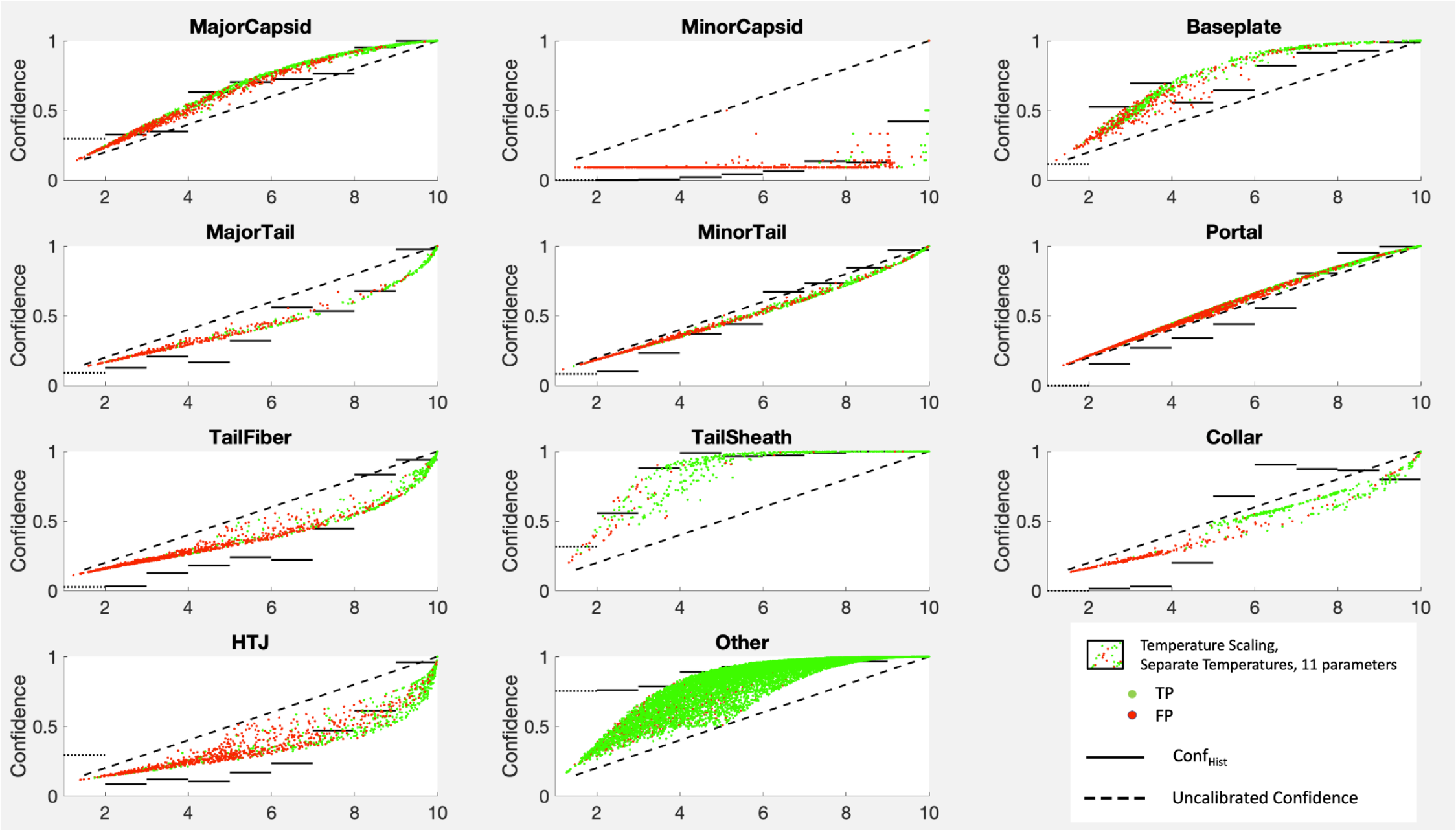}
\includegraphics[width=3.5in]{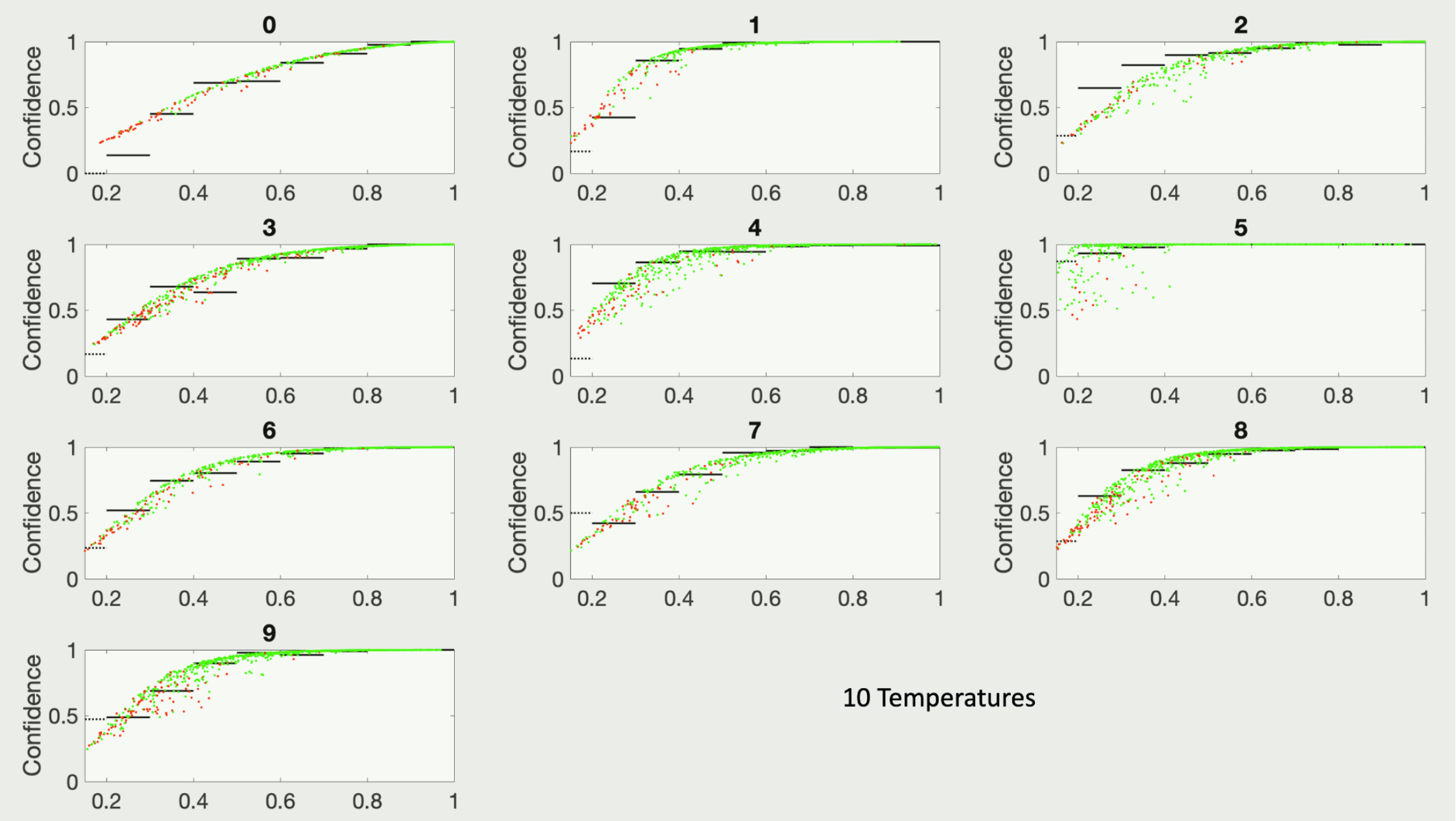}
\caption{Results for category-specific temperature scaling (CSTS) for PhANNs (top panel) and for MNIST (bottom panel). Note the convexity changes for TailFiber and HTJ, allowing much better fits than they got with one shared temperature. In each subplot, the horizontal axes show the score while the vertical axes show confidence measured as $y_k$.}
\label{fig_11}
\end{figure}

\section{Temperature Scaling with Awards (TSwA)}
A second modification is our attempt to exploit the additional information inherent in the choice of category by adding an “award” term to the score of the chosen category to account for the difference between the two curves in Figure 4. This results in
\begin{multline}
    \hat{Y}=\text{softmax}\Big(\frac{\log(y_1) + \delta_{1,k(X)}A_1}{T},\frac{\log(y_2) + \delta_{2,k(X)}A_2}{T}, \\
   \quad  ... \,, 
    \frac{\log(y_K) + \delta_{K,k(X)}A_K}{T}\Big)
\end{multline}
where $\delta_{ij}$ is the Kronecker delta, whose value is 1 if $i=j$ and otherwise 0. The $K+1$ vector $(A_1,A_2,...,A_K,T)$ is chosen to optimize the fit to the data for all categories at once. Thus it is the only technique we discuss that does NOT decompose the problem into $K$ separate fits. 

\begin{figure}[!t]
\centering
\includegraphics[width=3.5in]{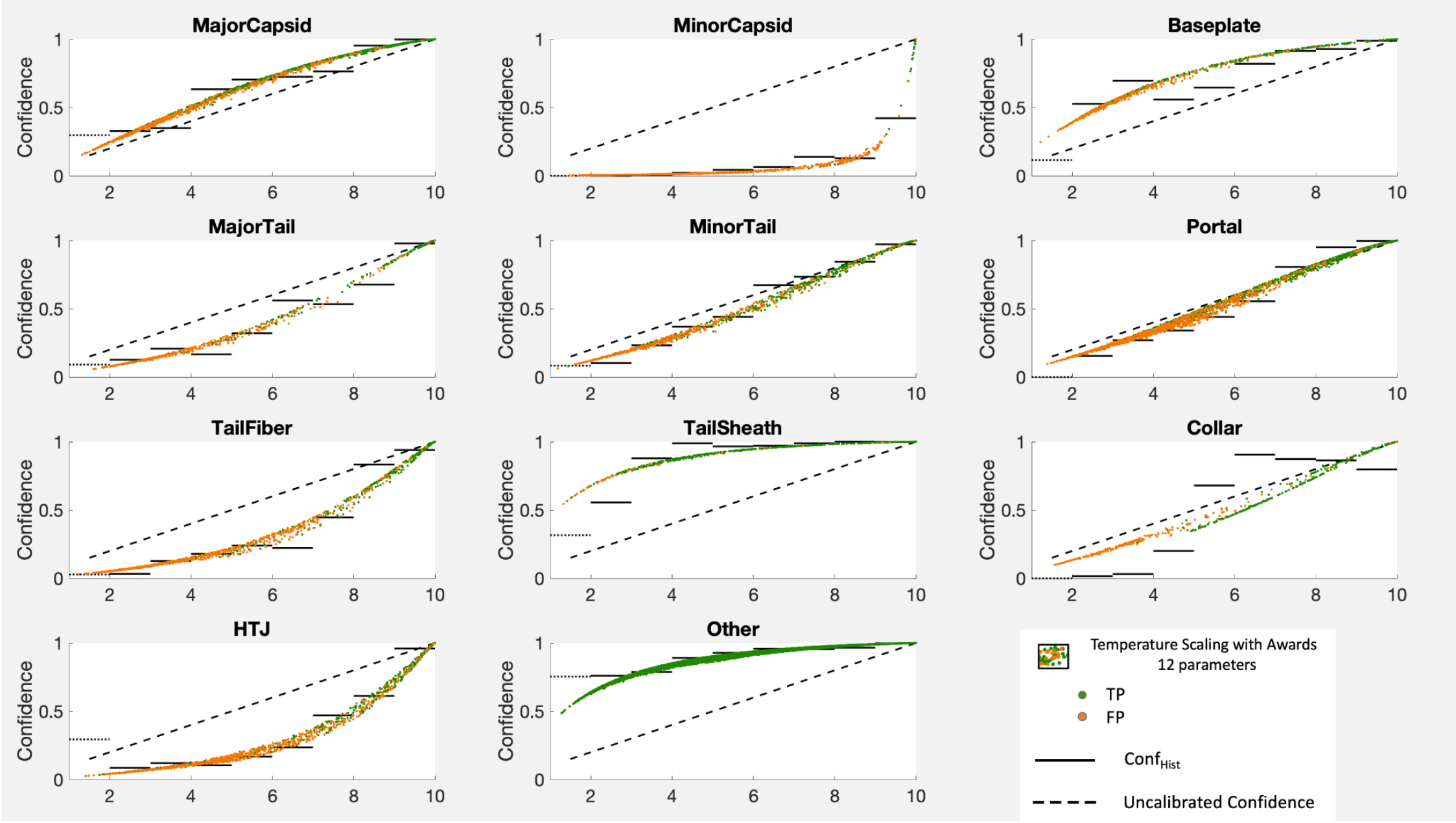}
\includegraphics[width=3.5in]{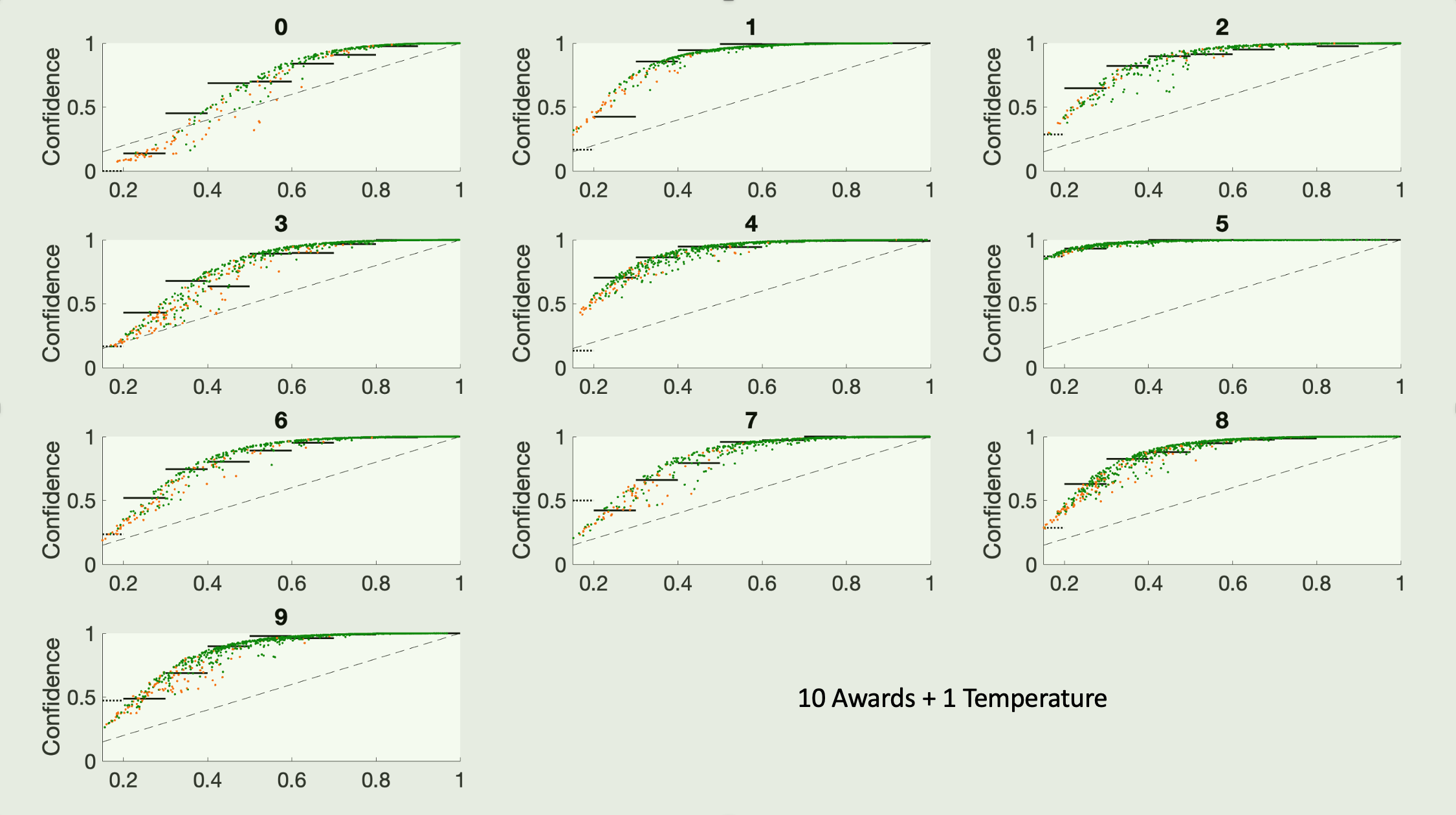}
\caption{Results for temperature scaling with awards (TSwA), again showing PhANNs in the top panel and MNIST in the bottom panel. Note the much tighter range of confidence values at each score, strikingly illustrated by the “Other” category. Note also the ability of $K$ awards with a single temperature to show the expressivity to capture not only the different convexities, but even resulting in a surprisingly good fit for “Minor Capsid”. In each subplot, the horizontal axes show the score while the vertical axes show confidence measured as $y_k$.}
\label{fig_12}
\end{figure}

We should mention that temperature scaling with awards can violate the first desiderata listed by Zhang et al\cite{zhang_mix-n-match_2020}. If any of our $A_i$ is less than zero, the winning category from $Y$ may no longer be the winning category from $\hat{Y}$. This means we have to separately store the value of the chosen category. Since we calibrate one category at a time, this does not seem an undue burden, especially in light of our incentive to award the winner. The purpose of $\hat{Y}$ is to have the correct confidence sitting in the $\hat{y}_{k(X)}$ entry, not to enable us to recalculate $k(X)$. 

Temperature scaling with awards is our favorite parametric technique, and outperforms CSTS on the PhANNs data (compare Figures 11 and 12). Notably either CSTS or TSwA performs much better than simple temperature scaling on PhANNs as well as MNIST (See appendix). While they admittedly need $K$ or $K+1$ parameters rather than 1, they are compatible with our top-label calibration philosophy, whereas simple temperature scaling is not.

It is possible to individualize both a temperature and an award to each category, resulting in a $2K$ parameter model. In the Appendix, we report on the resulting fits for MNIST. The performance is comparable to CSTS and TSwA. Given its doubling of the number of parameters without significant reduction in error, any criterion that penalizes a model for the number of parameters, e.g. AIC\cite{akaike_new_1974} or Occam’s Razor, would turn the $2K$ parameter model down in favor of the two models we presented above.

\section{Adjusting the bandwidth}
As a general rule, choosing the bandwidth for KDEs in a data-limited environment is more difficult than fitting temperatures or awards. KDEs suffer from the “trivial maximum at 0” problem\cite{guidoum_kernel_2020}. (The same is true for histograms. The smaller the bandwidth or the bin size, the better the fit to the data the kernels or the bins describe.) This is not a problem if a second test set is available, but that is a rare luxury for post-hoc calibrations on exploratory problems. 

The clue to a simpler, and much more stable way to choose the bandwidth comes from the early histogram binning literature in the form of an algorithm known as isotonic regression\cite{zadrozny_obtaining_2001}. This classic paper added constraints to the histogram method described above to force the resulting function Conf$_{{\rm Hist}}$ to be monotonic. The goal of monotonicity (or near monotonicity) was the key. 

The main failure mode for possible methods of choosing bandwidth is ending up with “wiggly” confidence curves, i.e. Conf$_{\rm{KDE}}$, with too many local minima. KDEs are sufficiently expressive to fit noise, leading to spurious “wiggles” in the resulting confidence curves (see Figure 13). Choosing the bandwidth by limiting the number of sign changes in the slope is a bulletproof, fast algorithm that can be easily implemented with no additional data. To facilitate our comparison of KDE based confidence estimators, we modify our notation to reflect the explicit $b$ dependence in the functions Conf$_{\rm{KDE}(b)}$.

\begin{figure}[!t]
\centering
\includegraphics[width=3.5in]{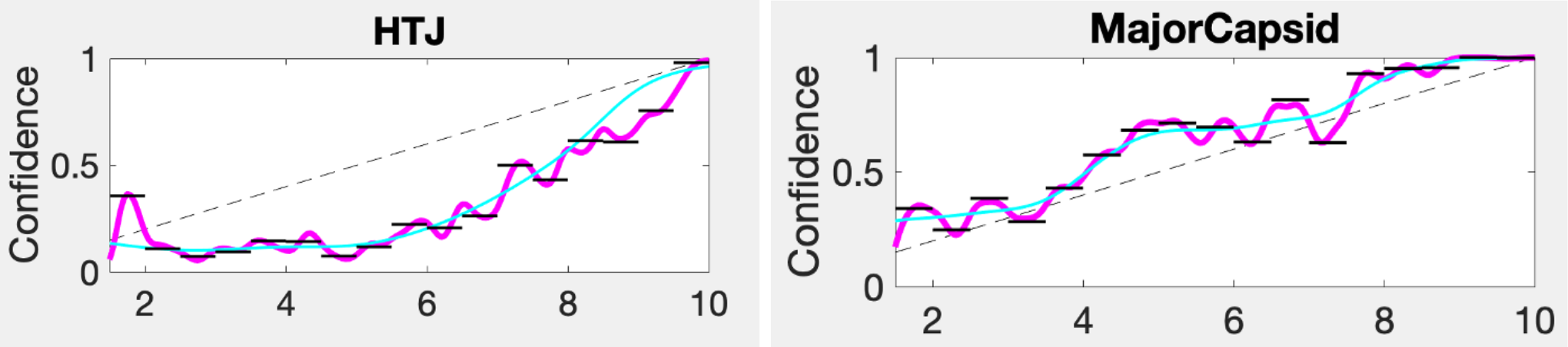}
\caption{The maroon curves shown are Conf$_{\rm{KDE}(b_{\rm{opt}})}$ curves using the optimal bandwidth from leave-one-out cross-validation fits to the data for the categories HTJ and Major Capsid. Too small a bandwidth always leads to wiggles that reflect the accidental (or real) clustering of the sample data in the test set. }
\label{fig_13}
\end{figure}

If we choose the number of sign changes to be zero, the corresponding bandwidth is
\begin{equation}
    b_{\rm{mon}} = \min \{b \,| \, \rm{Conf}_{\rm{KDE}(b)} \text{ is monotonic} \},	
\end{equation}

The Conf$_{\rm{KDE}(b_{\rm{mon}})}$ curves using category-specific $b_{\rm{mon}}$ for the PhANNs dataset are shown in Panel A of Figure 14. The fits are not bad, but can be improved significantly by relaxing the strict monotonicity requirements. Panel B shows the fits allowing one local minimum, i.e. using
\begin{equation}
    b_{\rm{mon2}} = \min \, \{ \, b \, \Big| \, \frac{d \,\text{Conf}_{\text{KDE}(b)}}{d\, \text{Score}} \quad \text{changes sign at most twice} \}
\end{equation}

\begin{figure}[!t]
\centering
\includegraphics[width=3.5in]{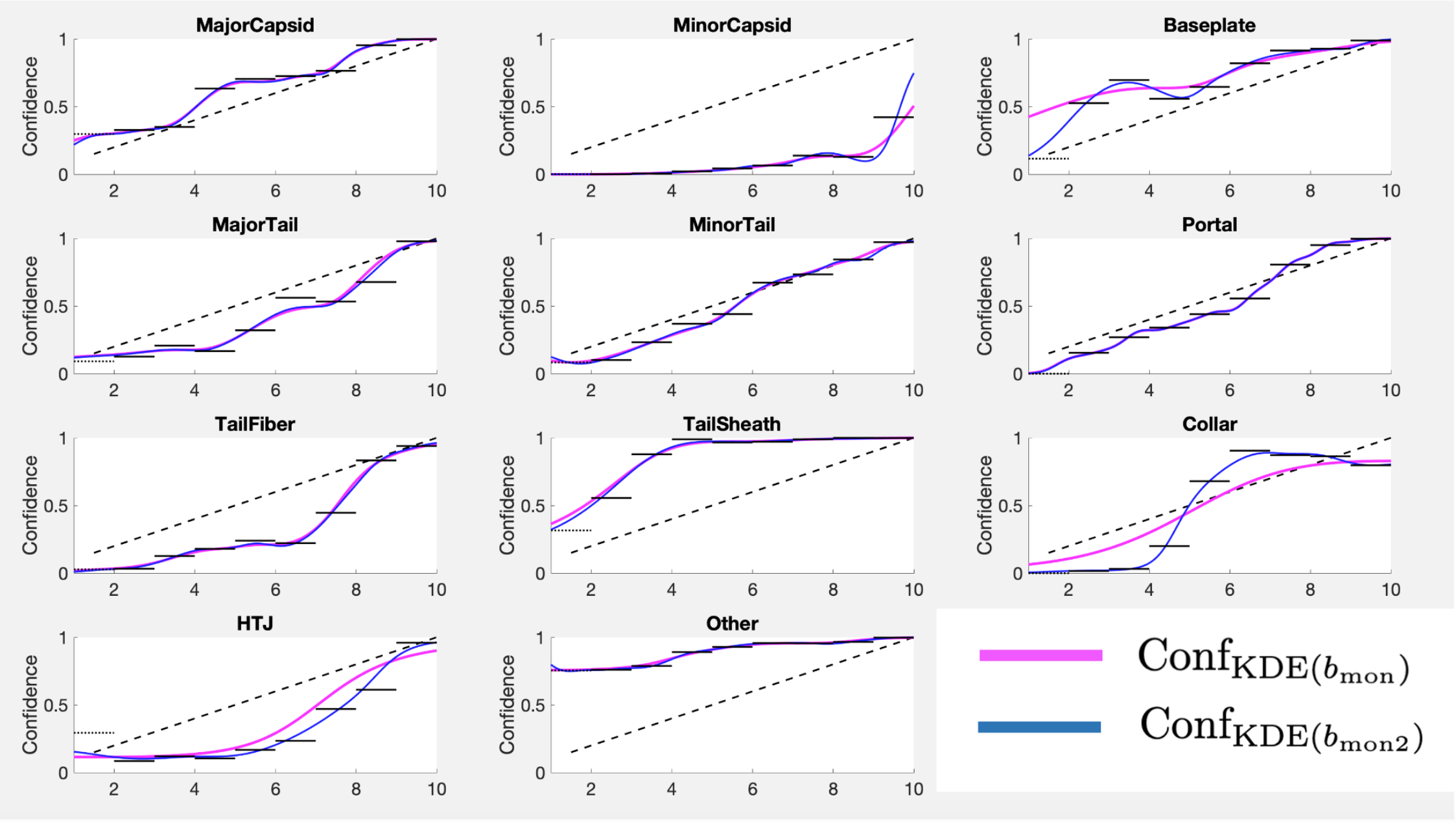}
\includegraphics[width=3.5in]{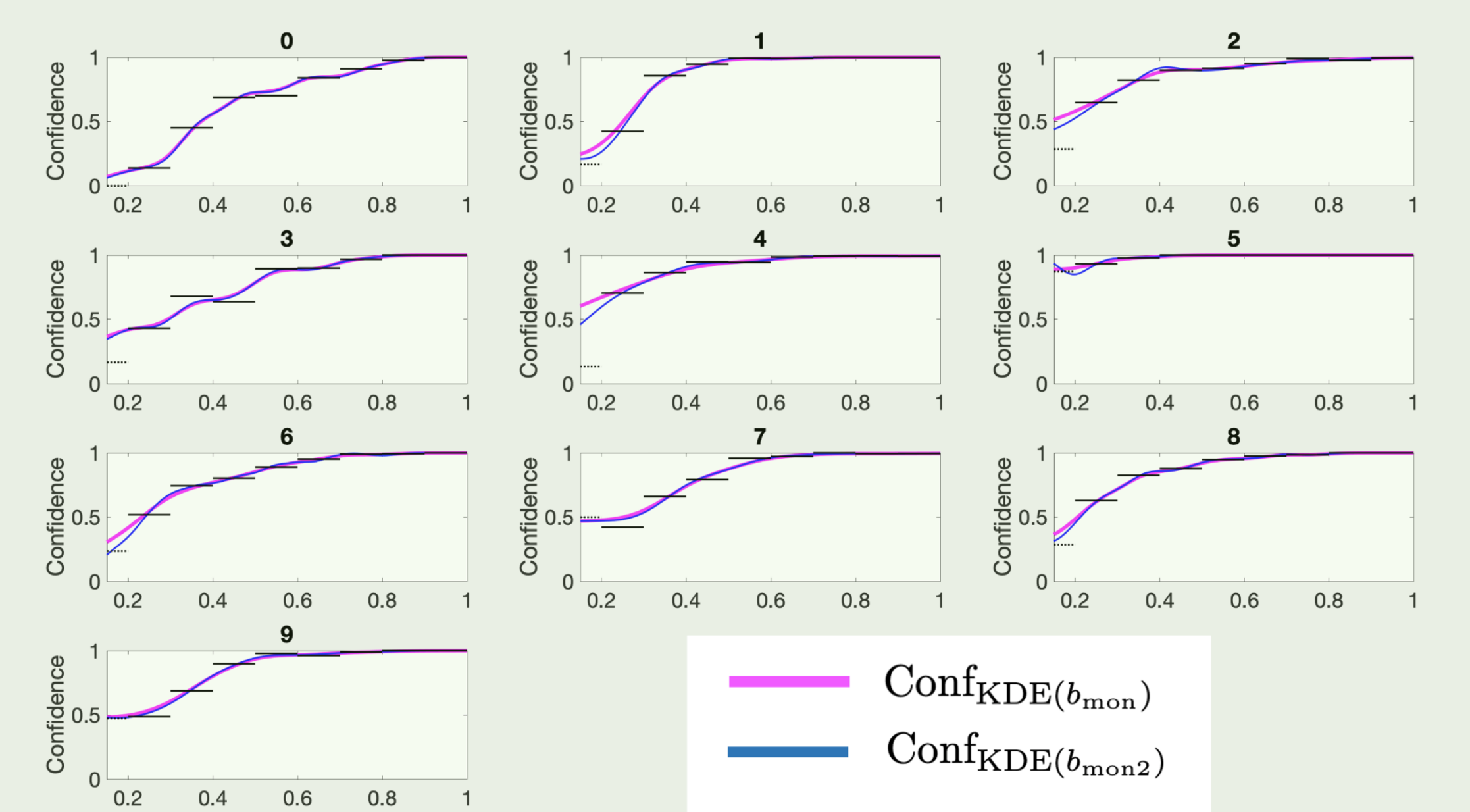}
\caption{Comparing monotonic performance of Conf$_{\rm{KDE}(b_{\rm{mon}})}$ (blue) and Conf$_{\rm{KDE}(b_{\rm{mon2}})}$ (magenta) for the PhANNs categories (panel A) and the MNIST categories(panel B).}
\label{fig_14}
\end{figure}

For most categories, allowing one local minimum changes the curve only slightly while improving the expressivity of the method immensely. Our recommendation is to use $b_{\rm{mon2}}$ for the bandwidth for both the TP and the FP estimates. The algorithm for finding $b_{\rm{mon2}}$ proceeds simply by starting at a low $b$, calculating the number of sign changes in $\Delta$Conf on a grid of $S$ values and incrementing $b$ until the number of sign changes reaches two or less. 

Using $b_{\rm{mon2}}$ improves generalization error over bmon. On the other hand, this is even more true for $b_{\rm{mon4}}$ and $b_{\rm{opt}}$
 (optimized with cross validation), but at the cost of many more wiggles and is achieved, in our opinion, by fitting the noise. For the more adventurous, exploring smaller bandwidths holds some rewards but is fraught with pitfalls. \emph{We would not recommend using any calibration without a visual check.}

\section{Discussion}
\subsection{Classification for Bioinformatics Problems}
In a sense, the application of machine learning algorithms to a problem is closely related to the application of compression methods to the same datasets. The difficulty of bioinformatic classification problems has been previously discussed in connection with the remarkable difference in lossless compressibility between text/images and biostrings\cite{wandelt_trends_2014,nalbantoglu_data_2009}. While standard compression tools (zip, rar, etc.) are commonly used to compress genome data, the level of compression these tools achieve when run on UNICODE encoded biostrings is 1:2 or 1:4, whereas bioinformatic specific compression can achieve lossless ratios of 1:1200 or more for higher organisms\cite{pavlichin_human_2013}. Furthermore, the compression algorithms employed are not similar, indicating that machine learning system performance may not generalize from image or text processing to bioinformatics processing, inasmuch as machine learning systems generalize across data using the same domain-specific structure that compression systems use to succinctly describe such data. The connection between compression and machine learning performance is seemingly well grounded, given that neural network architectures designed for natural language processing rather than compression are currently the best known method for compressing large text corpuses when runtime speed is not considered\cite{melis_state_2017}.

\subsection{Dataset composition frequencies}
During the course of this project, an additional trove of about 3000 new HTJ sequences became available for the PhANNs dataset. We added these additional samples to the 1227 previous HTJs in the test set and recalibrated. Figure 15 shows the results. The left panel shows the old and new estimated kernels. The first thing to note is that FP, and hence ${\cal K}_{\rm{FP}}$, is the same with or without the new samples. Recall that these are the false positives, those samples classified as HTJ whose true classes were not HTJ. Since every sample in the new trove has true class HTJ, the new trove of data had no effect on the false positive (red) distribution shown. This implies that confidence for HTJ (as defined in equation (6) or in equation (9)) can only go up by adding additional true HTJ samples. The lesson, though not new, is that the frequency of the representation of the different categories in the test set has a huge effect on the confidence scores. The bigger issue is: how well do the frequencies of category representation in the use-set mirror the frequencies in the test-set? 

\begin{figure}[!t]
\centering
\includegraphics[width=3.5in]{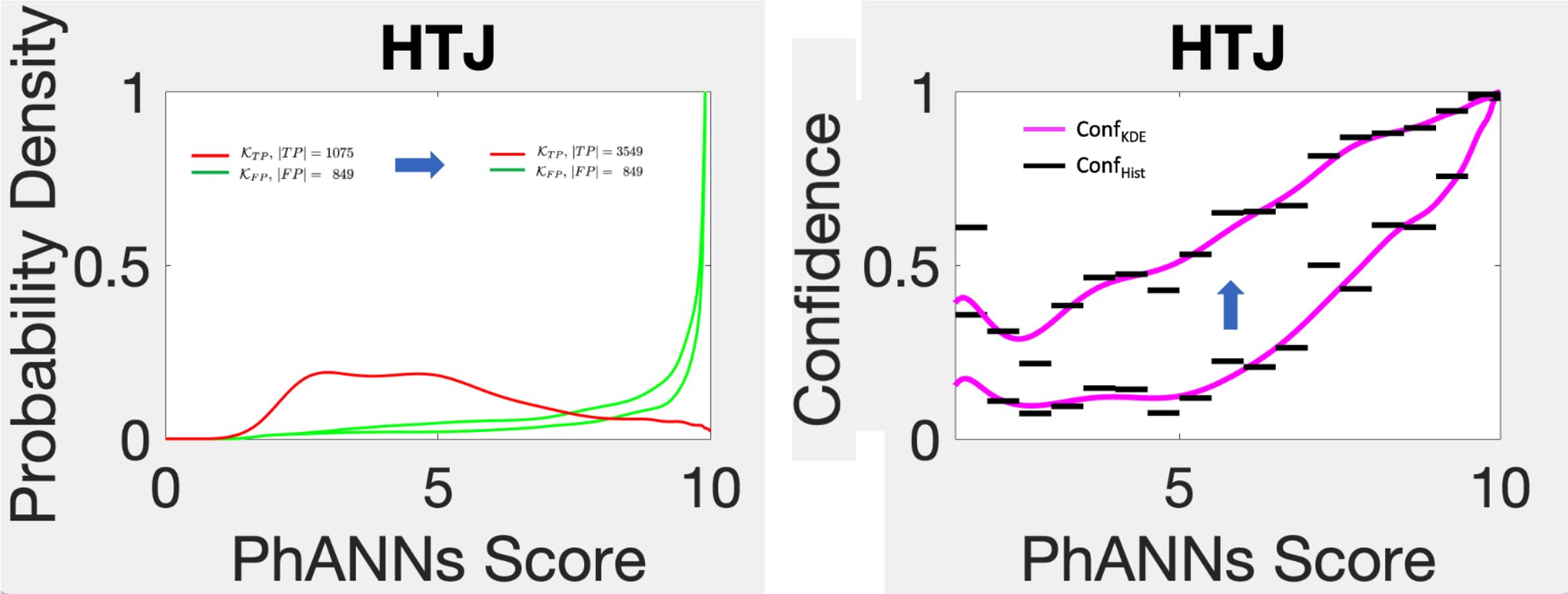}
\caption{The effect of adding an additional trove of “HTJ” samples to the test set used to predict confidence. Since this does not change FP, Conf$_{\rm{KDE}}$ can only  increase.}
\label{fig_15}
\end{figure}

The HTJ datatrove example serves as a broader reminder that the category representation in the test set has a huge effect on the confidence. Gupta and Ramdas have recently managed to push the quantitative assessment of error for histogram approaches to new levels. On the other hand, their analysis does not address the points made in the previous paragraphs: error from changes in the underlying distributions. In fact we believe the largest source of (unassessed) uncertainty is the mismatch between the frequency of the representations of the different categories in the test-set and its frequency in the use-set. This problem is called the label-shift problem\cite{podkopaev_distribution-free_2021}. With the rise of post-hoc calibration, an obvious suggestion is to keep use-statistics for the net and periodically recalibrate\cite{li_confidence-based_2006} with a test set that is curated to match the use-set statistics. For biological datasets in particular, this will be difficult to achieve.

This situation for PhANNs was exacerbated by (a reasonable) request from the journal (PLOS Computational Biology) that steps be taken to reduce the amount of interpolation rather than generalization. To this end, the ORFs being classified were de-replicated at 40\% and the remaining “seeds” were split up into the 12 groups, 11 for the cross-validation and one for the test set. When these seeds were expanded to include the de-replicated protein sequences, the size of the resulting groups varied widely and had a large effect on the representation of the different categories in the test set. This conflict between controlling representation frequency of the different categories and trying to keep easy interpolation answers out of the test set is likely to continue.

\subsection{Which densities?}
In the above presentation we introduced the two density estimators ${\cal K}_{\rm{TP}}$ and ${\cal K}_{\rm{FP}}$. It is of course equivalent to the more usual approach that estimates ${\cal K}_{\rm{TP}}$ and ${\cal K}_{\rm{P}}$, where
\begin{equation}
     {\cal K}_{P}  = \frac{{\cal K}_{\rm{TP}}|\rm{TP}| + {\cal K}_{\rm{FP}} |\rm{FP}|}{|\rm{TP}|+|\rm{FP}|}
\end{equation}

Our choice was predicated on the fact that separating out the false positives focuses on just how dependent our calculations are on the other categories and their representation in the test set. Note that ${\cal K}_{P}$ gives an alternate expression for the denominator in equation (9).

\subsection{The number of datapoints}
It came as somewhat of a surprise that the number of datapoints available for the calibration problem is not what one would naively expect. We got a lesson about this from the MNIST dataset. When examining calibration results for fully-trained nets (see Appendix), the results were surprising. Having most of the approximately 1000 points sitting at the score $S=1$ made the calibration effort nearly impossible. For example, for category “0”, there were only four TP scores less than 0.99 and eleven FP scores total. This is far too sparse to assess score dependence. On the other hand, it is not a likely scenario for the exploratory problem-type we focus on in this work. We specifically included the comparison to MNIST in order to underline such features.

In Figure 16 we show the fully-trained MNIST results in the form of the joint distribution and the resulting calibration curves using kernel densities with $b_{\rm{mon2}}$. A reasonable confidence estimation approach in this case is to just stick to confidence given only the category, as in equation (1). If a top-label approach is required, Conf$_{\rm{KDE}(b_{\rm{mon})}}$ produces acceptable results as does Conf$_{\rm{CUM}*}$, though most of the resulting optimal cutoffs $\theta_{\rm{opt}}$ fall between 0.99 and 1.00.
\begin{figure}[!t]
\centering
\includegraphics[width=3.5in]{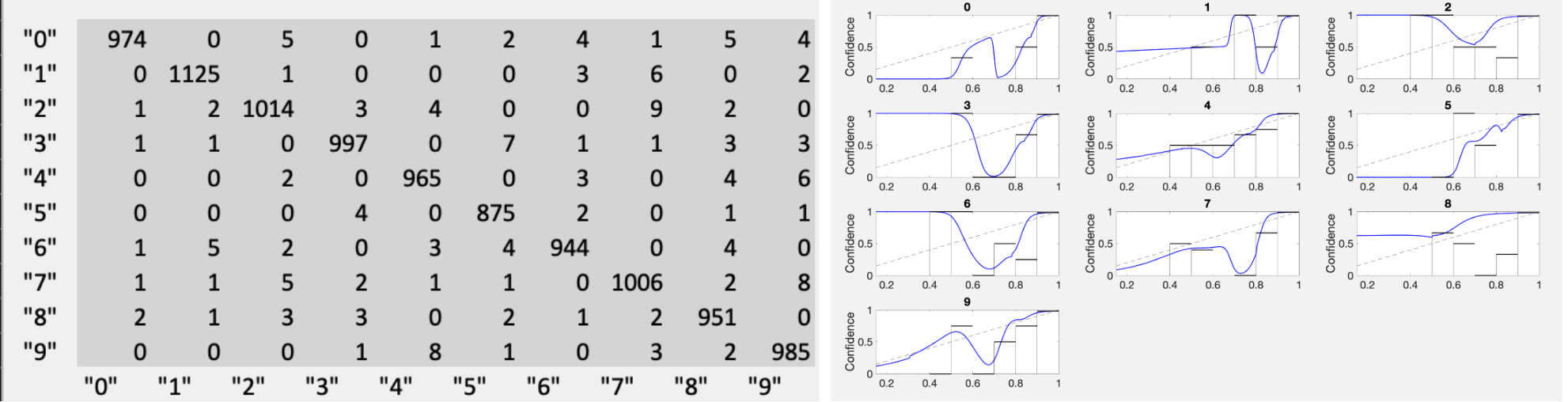}
\caption{The joint distribution from the “Fully trained” MNIST results. Note the very few erroneously classified samples shown here as the counts in off-diagonal entries. The right panel shows the blind application of our recommended Conf$_{\rm{KDE}(b_{\rm{mon2})}}$ and provides a warning against using any calibration without a visual check.}
\label{fig_16}
\end{figure}

Sparse data is an issue for confidence calibration more generally. Typically, this happens in two regions: high scores for false positives and low scores for true positives. Local sparsity was also the most common reason for our Conf$_{\rm{KDE}}$ curves to develop spurious wiggles. Our better bandwidth selection scheme gets around most of these problems but some persist. At scores well below the first datapoint, the tails of the different normal distributions added up can make the kernel density’s imagination go wild. This is the reason our plots begin at PhANNs score 1.5 (MNIST 0.15) and the lowest histogram bar is shown dashed in all of our figures. 

\subsection{Boundary Correction}
Kernel densities on intervals with a boundary do not count quite the way we would like since some of the probability represented in the distribution falls outside the interval. While at overly-small bandwidths this effect can be quite pronounced, at any reasonable bandwidth the effect is minimal and can be ignored.

\section{Comparing Methods}
The following chart compares the summary performance of our new methods to histograms, shared temperature scaling (STS) and uncalibrated networks. Our first observation is that the uncalibrated nets, shown by the last columns in each category, do well for some — notably MinorTail and Portal — but lag significantly in other categories, corroborating our assertion above that post-hoc calibration is always a good idea. Our second observation is that sharing a temperature (STS, in yellow) is not a good idea for the disparate categories typical of exploratory problems, although it performs quite well for MNIST (see Figure 19 in the Appendix). While histograms of course work very well in all cases\cite{gupta_top-label_2022,gupta_distribution-free_2020, gupta_distribution-free_2021},  our new techniques work comparably well or better without the noise sensitivity that have made most users shun this method. The chart also shows why we have been featuring KDE methods, especially KDE($b_{\rm{mon2}})$, which generally work as well or better. The difference between KDE($b_{\rm{mon}})$ and KDE($b_{\rm{mon2}})$ is small and in our opinion worthwhile, but as we argue in section 12 and the Appendix, which wiggles are real is not a simple call. KDE($b_{\rm{mon}})$ does not lose out to KDE($b_{\rm{mon2}})$ by enough to look significant in such aggregate comparison. The detailed visual comparisons in section 11 are our strongest argument, although that comparison relies on the human eye providing some pattern recognition. Category-specific temperature scaling (CSTS, light blue) performs surprisingly well. Again, our best argument in favor of TSwA over CSTS is through a human eye comparing the fits in Figures 11 and 12. Finally, cumulative confidence with an optimal cutoff (CwOC, dark blue) is competitive but not stellar, just as we noted in section 7.

\begin{figure}[!t]
\centering
\includegraphics[width=3.5in]{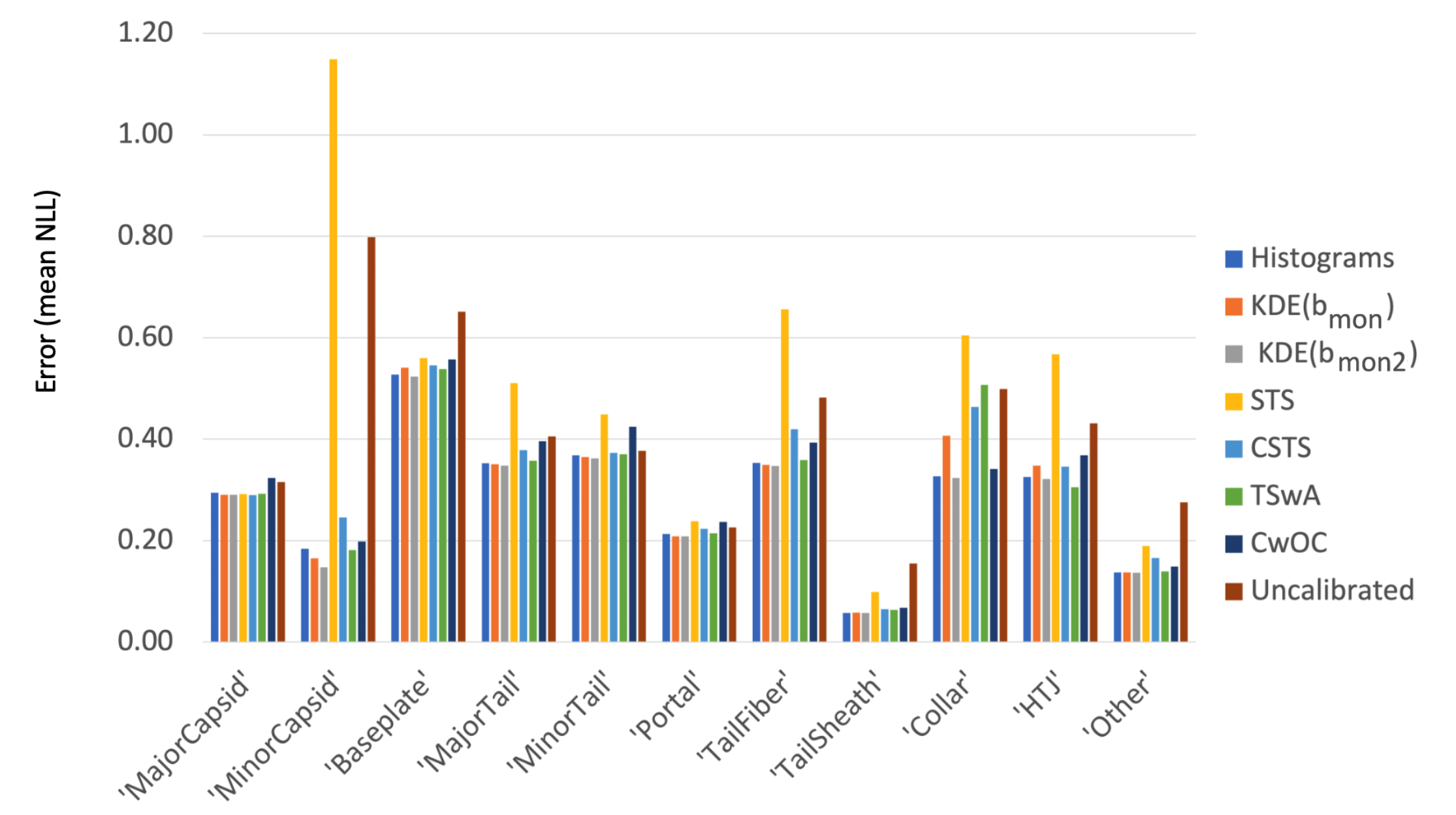}
\caption{The errors for confidence calibration methods compared on PhANNs categories. The errors shown are the average negative log likelihood per sample. The techniques compared are discussed in the sections above: histograms (section 4), kernel density based estimators (sections 5 and 12), shared temperature scaling (STS, section 9), category specific temperature scaling (CSTS, section 10), temperature scaling with awards (TSwA, section 11) and cumulative confidence with optimal cutoff (CwOC, section 7). The uncalibrated scores use the $y_{k(X)}$ entry from the MLS output $Y$.}
\label{fig_17}
\end{figure}

\section{Conclusions}
In the foregoing manuscript, we describe several new techniques for post-hoc calibration of machine learning systems. Specifically, we aim at a problem size and imbalance typical of exploratory problems – machine learning problems exploring new categories whose very definition depends on a dialog between the MLS and the researchers. We further argue that for such problems, top-label calibration makes better sense than other approaches.

Our favorite technique for such problems is kernel density estimator-based confidence, Conf$_{\rm{KDE}(b_{mon2})}$, which comes with a new robust algorithm for bandwidth selection. We suspect that it was exactly the lack of such a robust bandwidth selection algorithm that has limited the prior use of kernel density estimates for confidence calibration. We also introduced three new top-label oriented parametric techniques. The first of these models (CSTS) is a straightforward application of temperature scaling to the $K$ one-versus-all problems that result from breaking up the multi-classification into the top-label versions. The second (TSwA) directly attacks the difference between determining confidence with or without the information provided by the predicted category. $K$ of the $K+1$ parameters in TSwA are “awards” which are added only to the predicted category’s score before the otherwise standard temperature scaling is applied to the shifted scores. The idea is to modify the initial score to account for the information contained in the MLS selecting a specific category. TSwA performs visibly better than CSTS on PhANNs (compare Figures 11 and 12). Again, we recommend a visual check for any calibration method.

Calibration of confidence, like calibration of accuracy, must be post-hoc to prevent bias. Parametric approaches, with parameters trained from the test set used for the calibration, appear to compromise this benefit. Using only a few generic parameters keeps this bias to a minimum. We use the MNIST problem as a way to estimate how large a bias our generic parameters introduce and find that, at the problem size typical for exploratory problems, the sample noise swamps the bias. Our belief is that the largest source of confidence error comes from yet a different source: the mismatch between the test set used in the calibration and the use-set, i.e. the field use composition of categories. Compared to such "label-shift" error\cite{podkopaev_distribution-free_2021}, the differences between the techniques presented are small. The size of the sample and uncertainty in the use-set frequencies of the different categories leave us with a limiting knowledge gap: better calibration is not available without better data. 

Our experience with the fully-trained MNIST example highlights the uniqueness of datasets for exploratory problems in contrast to relatively large and balanced datasets that represent “already solved“ problems. “Already solved” problems have been the focus of much of the confidence calibration literature and explains why that literature ignores the obvious difference between top-label confidence and approaches that ignore the category predicted. The distinction between the two approaches is moot if all scores are nearly zero or nearly one.

Exploratory problems represent a type of problem that will be around for the foreseeable future. We hope this collection of calibration techniques and lessons will be useful to the community.

\section*{Acknowledgments}
We are grateful for discussions with Barbara Bailey, James Darwin Nulton, Katelyn McNair and the Biomath group at SDSU. We especially benefited from Barbara Bailey’s repeated references to the eye$^2$ method over many years of discussions. We thank A. Ramdas for helpful comments based on an early draft of the manuscript

{\appendix[MNIST Training Details]
During the early stages of algorithm research, it is not uncommon to see members of different research groups sharing a preference for a given algorithm. Some of this comes down to subjective preference and local fashion, but a second source of preference differences can come from testing algorithms on different data. We would not be surprised to learn that different algorithms can more or less closely match the calibration curves for some problem domains.

To combat the risk of “domain bias,” and thereby temper our group’s biased background largely dealing with biological datasets, we created ensembles of MNIST classifiers. Additionally, to minimize the chance of sampling bias, we created a system which allowed us to compare algorithm performance differences across thousands of runs.

\subsection*{MNIST dataset preparation}
The MNIST dataset comes as 60,000 training examples and 10,000 testing examples. Every example is a pair $(X, \rm{TrueClass}(X))$, where $X$ is a 28x28 grayscale image, $X \in~R^{784}$, and TrueClass is a categorical label, 
TrueClass~$\in \{0, 1, 2, 3, 4, 5, 6, 7, 8, 9\}$.

To prepare our data for ensemble member training, we split the test data into a 50,000 element “training set” group, and reserved 10,000 elements as “test set 1”. The MNIST-provided test set was dubbed “test set 2”.

\subsection*{Network training}

\noindent
\textbf{Minibatch selection:} Mini batches of 1,000 elements were sampled, with replacement, from the “training set” group.

\vspace{6pt}

\noindent
\textbf{Architecture:} All neural networks were run in stages: 
\begin{enumerate}
    \item the input data was flattened into a 784 element vector,
    \item the output of 1 was run through a fully connected ReLU layer to 300 outputs,
    \item the output of 2 was run through a fully connected ReLU layer to 100 outputs,
    \item the output of 3 was run through a 10 element softmax layer.
\end{enumerate}
We denote the resulting output as $Y(X)$.

\vspace{6pt}
\noindent
\textbf{Loss function:} We minimized categorical cross entropy, namely 
\begin{equation}
    \nonumber
    -\sum_X(\rm{onehot(TrueClass}(X))*\log(Y(X))) 
\end{equation}
using the ADAM optimizer\cite{kingma_adam_2014}, with a lambda of 1e-3.

\vspace{6pt}
\noindent
\textbf{Training steps:} For under-trained networks, we trained for 400 eval-gradient-update steps. Normal accuracy for an under-trained network, measured using “test set 2,” was approximately 88\%. Under-trained networks were used to more closely mimic the available data for our exploratory bioinformatics problem. For the fully-trained (over-trained) networks, we trained for 10,000 eval-gradient-update steps. Normal accuracy for an over-trained network, measured using “test set 2,” was approximately 99.7\%

\subsection*{Ensemble Sampling}
We trained 1,000 networks as described above. When a new ensemble output was desired, we sampled from the $\genfrac(){0pt}{2}{1000}{10}$
possible combinations of 10 members from those 1,000 ensembles, ran each network on both datasets, then took the softmax of their summed log outputs. This gave us a fresh pair of test sets on which to compare calibration algorithms.

\subsection*{Calibration method tuning}
As all of the calibration methods tested had one or more variables that needed tuning to data, and generalization performance was of principal concern. We tuned each calibration method to minimize NLL on the “ensemble test set 1”, then measured the calibration errors ECE1, ECE2, and NLL on both the “ensemble test set 1” and “ensemble test set 2” datasets for reporting. 

The temperature scaling methods examined with the above double-test-set design were: single temperature scaling (STS), categorical temperature scaling (CSTS), single temperature scaling with categorical awards (TSwA), and categorical temperature scaling with categorical awards (CSTSwA, 20 parameter model).

The bandwidth selection methods examined with the above double-test-set design were: categorical monotonic bandwidths $(b_{\rm{mon}})$, categorical bandwidths allowing one local minimum $(b_{\rm{mon2}})$, categorical bandwidths allowing two local minima $(b_{\rm{mon4}})$ and optimized categorical bandwidth $(b_{\rm{opt}})$; the latter were selected by using leave-one-out cross validation .

\subsection*{MNIST results}
The figures below show the median of 1000 MNIST ensembles with parameters that were trained on testset 1 and applied to both testset 1 and testset 2. In the figures, each method is associated with a color. The left column of each color always represents the performance on testset 1, the right its performance on testset 2. If significant bias were present from the parameter selection, we would expect that performance on testset 2 would be uniformly worse. Instead, the differences appear random (about as many higher as lower). This is evidence of a lack of bias large enough to dominate the error due to sampling, whose magnitude is revealed in the figures through these left-right differences.

Figure 18 shows the surprisingly similar performance of our four temperature scaling related methods with the notable exception of STS (red), which performs significantly worse for category “5”. As discussed in section 9, simple temperature scaling works very well on examples like MNIST where all the categories need almost the same temperature. The other three models perform almost identically. We see this as arguments in favor of our 10 and 11 parameter models (CSTS and TSwA respectively) but not the 20 parameter model (shown in blue), whose spurious generalization risk due to the extra parameters is not justified without significant improvement in performance\cite{akaike_new_1974}, as discussed at the end of section 11 in the main text.

\begin{figure}[!t]
\centering
\includegraphics[width=3.5in]{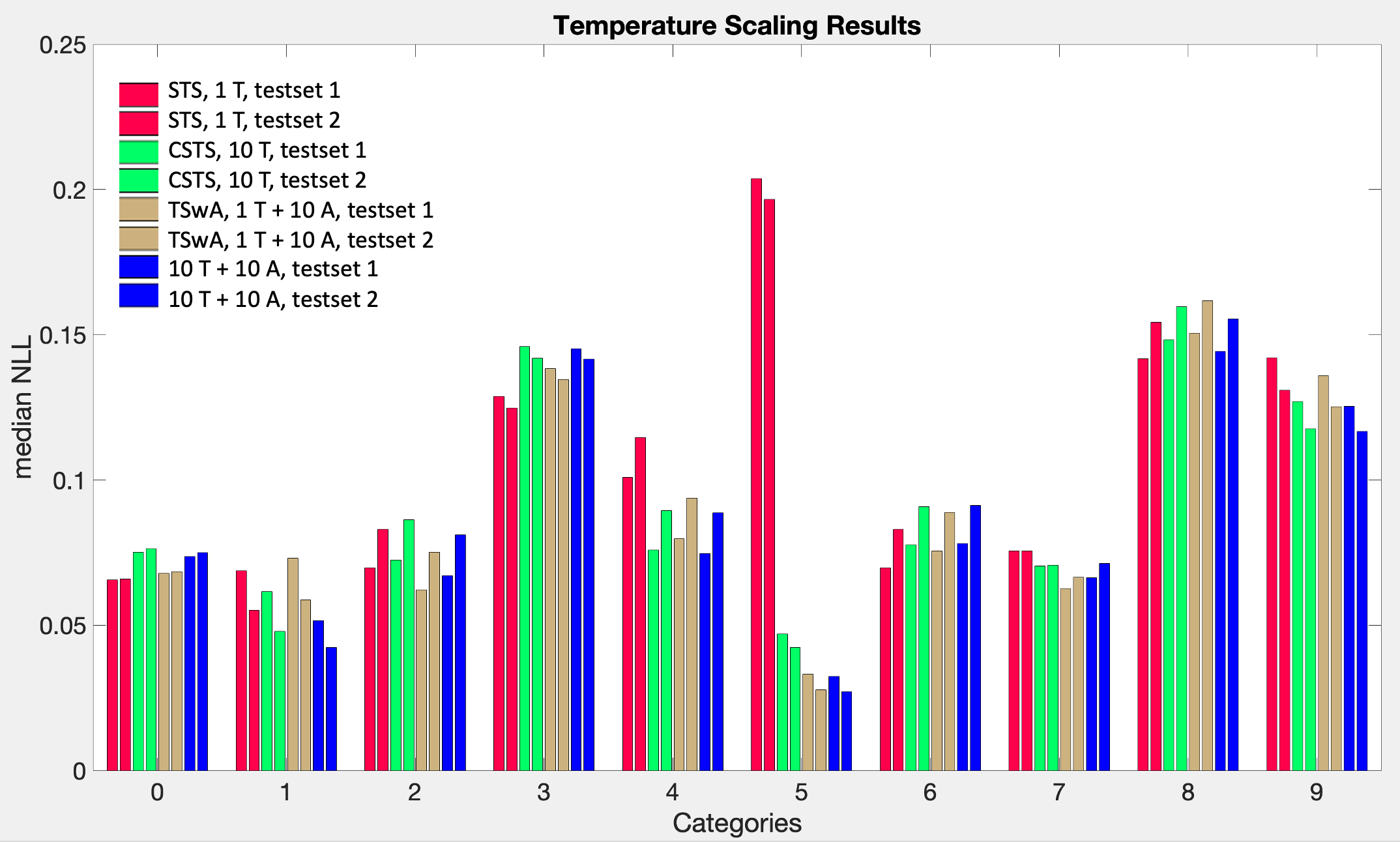}  
\caption{The performance of temperature scaling related methods applied to MNIST. All but the left method (red) represent an adaptation of temperature scaling to top-label calibration. }
\label{fig_18}
\end{figure}

The MNIST bandwidth selection results shown in Figure 19 are pretty much as expected, with larger bandwidths forcing larger NLL. By definition, the bandwidths $b_{\rm{mon}} > b_{\rm{mon2}} > b_{\rm{mon4}}$ represent progressively weaker constraints on the number of sign changes in the slope of the confidence curve, with the less constrained cases (bigger $N$ in $b_{\rm{monN}}$) producing smaller NLL values.

Recall however, that choosing the bandwidth by optimizing the fit with leave-one-out cross-validation on the PhANNs dataset almost always gave unacceptably wiggly Conf$_{\rm{KDE}(b_{\rm{opt}})}$ estimates (See Figure 13).
Local pockets of data sparsity or excess in either TP or FP can combine with the  “trivial maximum at zero” problem\cite{guidoum_kernel_2020}, to choose wiggly confidence curves. 
We believe this instability has kept KDEs from wider use for confidence estimation. This instability is a serious issue for the ``exploratory problems'' PhANNs represents though  a minor issue for the well balanced problems MNIST represents. The $b_{\rm{mon}}$ and $b_{\rm{mon2}}$ fits in Figure 14 are almost identical for the MNIST data. In fact for the MNIST examples we see $b_{opt}$ values even smaller than $b_{\rm{mon4}}$ on the average. As discussed in section 12, our recommendation is to use $b_{\rm{mon2}}$ (green in Figure 19) to avoid overly wiggly confidence curves that fit mostly noise. Note that we again see evidence that the sample noise is significantly bigger than any bias from the training; else we would see uniformly worse performance on testset 2.

\begin{figure}[!t]
\centering
\includegraphics[width=3.5in]{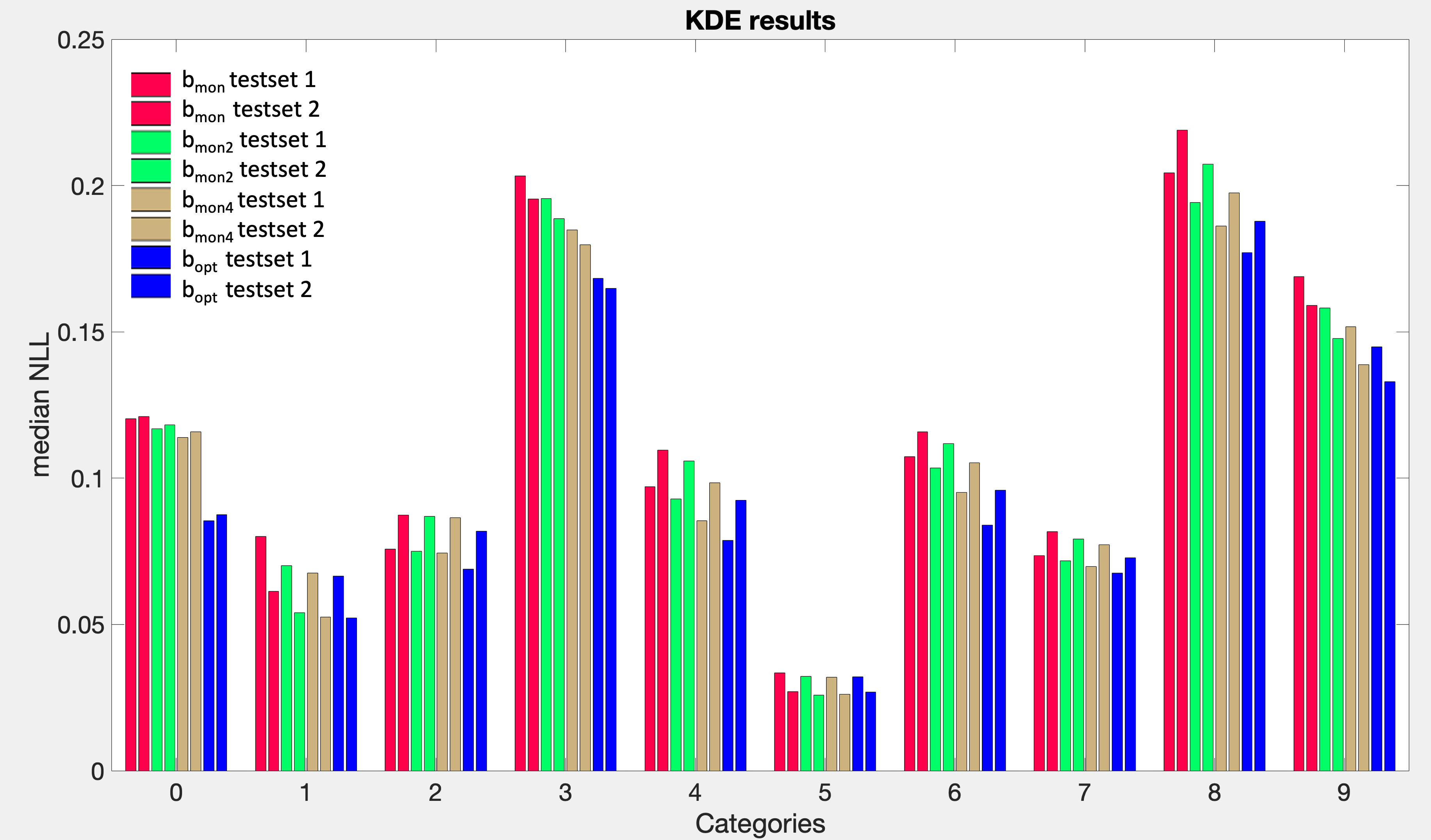} 
\caption{The effect of bandwidth selection on median negative NLL for the MNIST data. The testset1 results are always shown to the left of the testset2 results.}
\label{fig_19}
\end{figure}

}

\bibliography{bibtex/bib/anca}
\bibliographystyle{IEEEtran} 

\section{Biography Section}


\vspace{-133pt}

\begin{IEEEbiography}[{\includegraphics[width=1in,height=1.25in,clip,keepaspectratio]{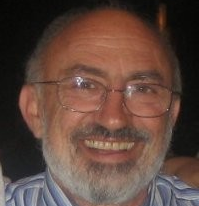}}]{Peter Salamon}
 received his PhD in chemical physics from the University of Chicago, He has been a professor in the Mathematics Department at San Diego State University from 1980 to the present, with visiting positions at various institutions including the Hebrew University in Jerusalem, the University of Heidelberg, and the University of Copenhagen. He is known for his work in finite-time thermodynamics and for having introduced the use of ensembles into neural networks. 
\end{IEEEbiography}

\vskip -2\baselineskip plus -1fil

\begin{IEEEbiography}[{\includegraphics[width=1in,height=1.25in,clip,keepaspectratio]{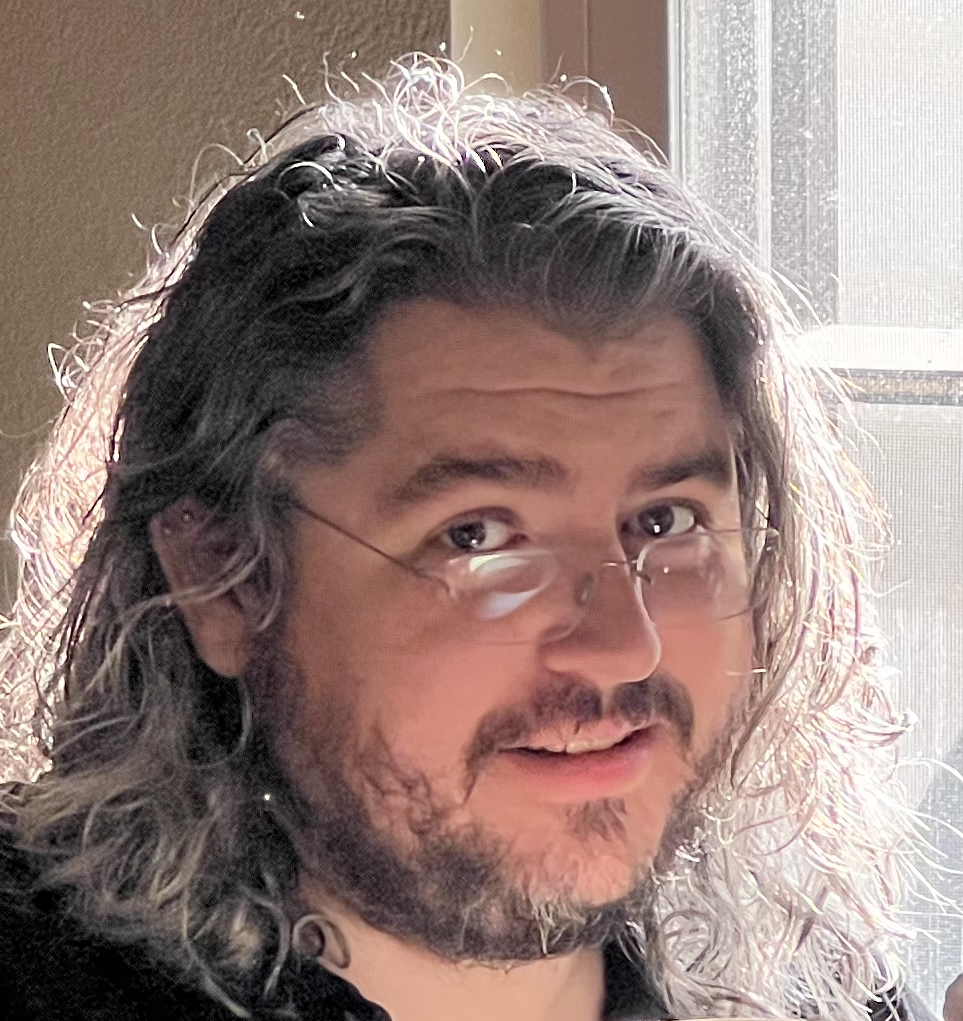}}]{David Salamon}
studied computer science at San Diego State University, the University of California Santa Barbara, and St. John's College. He was an early programmer at ProCore (NYSE: PCOR), WealthFront (NYSE: UBS), and Apptimize (YC S13.) He's interested in machine learning, distributed systems, and prosocial technology.
\end{IEEEbiography}

\vskip -2\baselineskip plus -1fil

\begin{IEEEbiography}[{\includegraphics[width=1in,height=1.25in,clip,keepaspectratio]{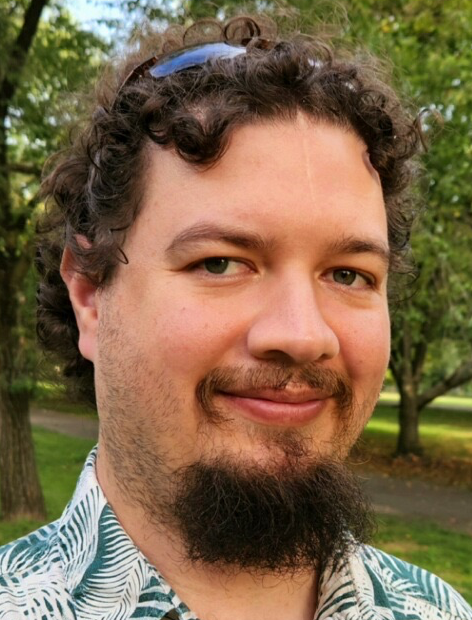}}]{V. Adrian Cantu}
obtained his BS from the Centro de Ciencias Genomicas (genomic sciences center) at UNAM (National University of Mexico), and a Masters in Biochemistry from the Instituto de Biotecnologia at UNAM. He received his PhD in 2020 from the SDSU / Claremont Graduate Institution Joint Doctoral program, applying machine learning approaches to predict the function of unknown genes encoded by bacteriophages (viruses that infect bacteria). He is currently a postdoctoral fellow at the University of Pennsylvania School of Medicine, and uses bioinformatics to investigate the insertion sites of retroviruses in mammalian genomes and the molecular consequences of gene therapy.
\end{IEEEbiography}

\vskip -2\baselineskip plus -1fil

\begin{IEEEbiography}[{\includegraphics[width=1in,height=1.25in,clip,keepaspectratio]{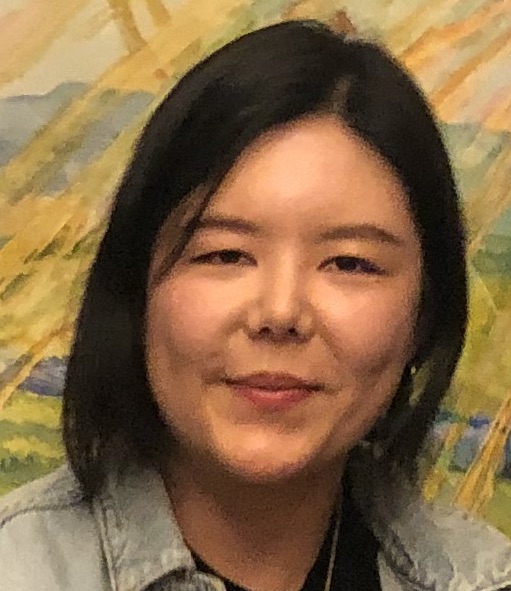}}]{Michelle An}
earned her B.S. in General Biology from the University of California, San Diego and her M.S. in Bioinformatics and Medical Informatics from San Diego State University. Her thesis work focused on developing and applying bioinformatics tools and machine learning-based analyses of bacterial and bacteriophage genomes. Since 2021 she works as a Bioinformatics Engineer at Helix Inc.
\end{IEEEbiography}

\vskip -2\baselineskip plus -1fil

\begin{IEEEbiography}[{\includegraphics[width=1in,height=1.25in,clip,keepaspectratio]{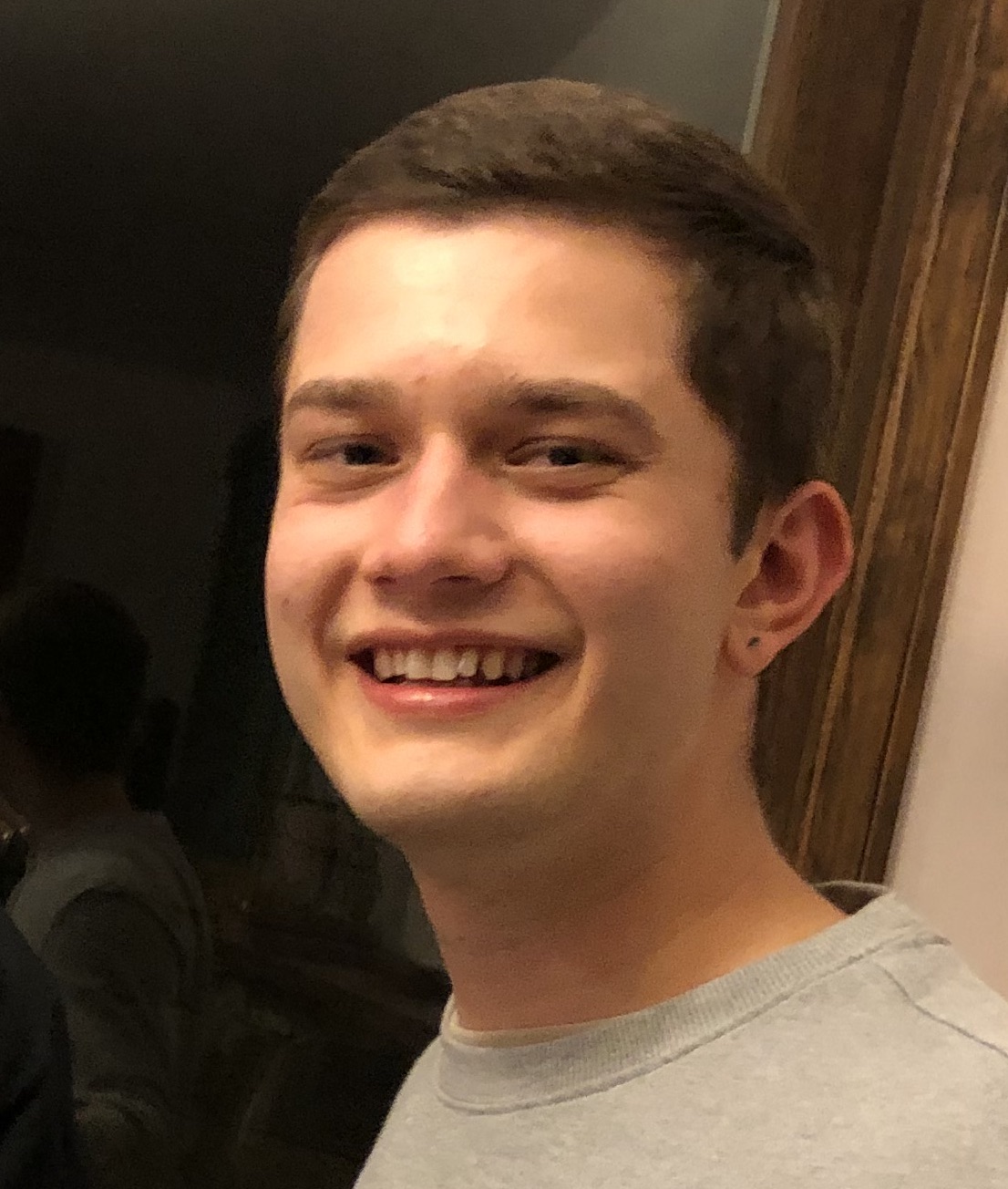}}]{Tyler Perry}
obtained his BS in Computer Science in 2021 from San Diego State University, and was involved in applying embeddings to augment the performance of artificial neural networks to classify and predict the function of unknown proteins encoded by bacterial viruses. On graduating, Tyler joined Google as a software engineer.
\end{IEEEbiography}

\vskip -2\baselineskip plus -1fil

\begin{IEEEbiography}[{\includegraphics[width=1in,height=1.25in,clip,keepaspectratio]{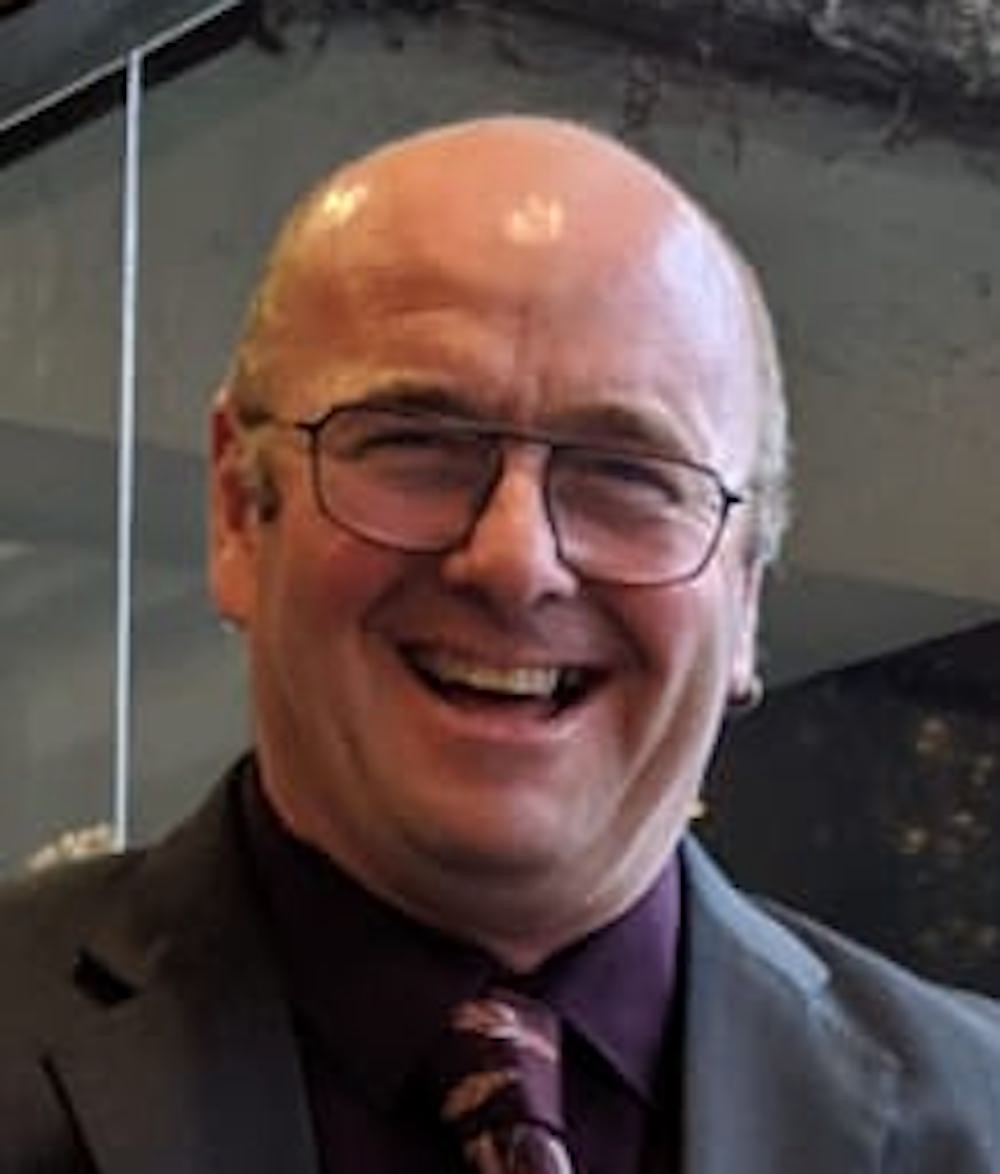}}]{Robert (Rob) A. Edwards}
earned his PhD in Microbiology (1994) from the University of Sussex, Brighton, UK, then moved to the US for postdocs investigating host-pathogen interactions, first at the University of Pennsylvania and subsequently at the University of Illinois at Urbana Champaign. After spending four years as faculty at the University of Memphis, he was a consultant with the Fellowship for Interpretation of Genomes, then joined the faculties of the Biology and the Computer Sciences departments at San Diego State University and served as adjunct faculty at the The Burnham Institute for Medical Research. He was a visiting scientist at Argonne National Labs and a visiting scholar at the Federal University of Rio de Janeiro. In 2020, he moved to Adelaide, Australia to join the College of Science and Engineering at Flinders University. His research interests encompass bioinformatic, genetic, and genomic analyses of bacteria and bacteriophages, and develops computational tools for analysis of environmental microbiology. He is a fellow of the American Academy of Microbiology and a founding member of the Viral Information Institute.
\end{IEEEbiography}

\vskip -2\baselineskip plus -1fil

\begin{IEEEbiography}[{\includegraphics[width=1in,height=1.25in,clip,keepaspectratio]{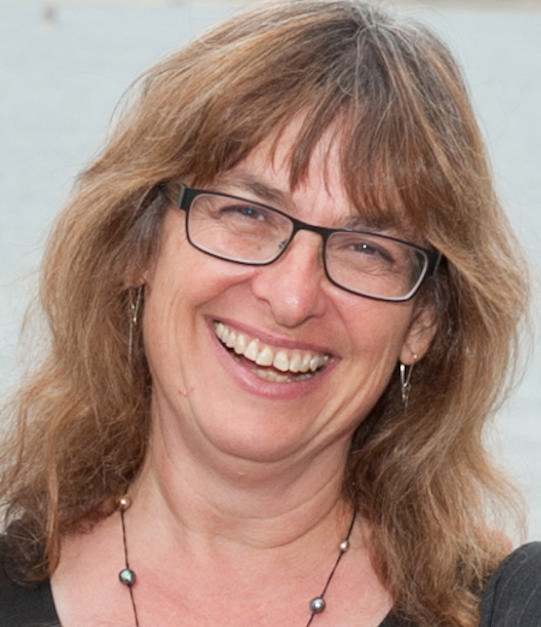}}]{Anca Segall}
obtained her PhD in Genetics from the University of Utah (1987), then was a visiting scientist at E. I. du Pont Co., Wilmington DE, and a postdoctoral fellow and National Research Council fellow at the NIH. In 1994 she joined the faculty of the Department of Biology at San Diego State University and is former chair of that department. She is also on the faculty of the Chemistry and Biochemistry Department, and member of the Computational Science Research Center.  Her research centers on the functions of bacteriophages in environmental and human microbiomes, mechanisms of recombination and horizontal gene transfer, and phage therapy applications to combat antibiotic resistance, and employs biochemistry, genetics, genomics, and computational approaches. In 2014 she founded the Viral Information Institute at SDSU, and is a fellow of the American Academy of Microbiology.
\end{IEEEbiography}

\vfill

\end{document}